\documentclass[]{fairmeta}
\newcommand{\dataset}{\ensuremath{\mathcal{D}}}
\newcommand{\teacher}{\ensuremath{M}_{teacher}}
\newcommand{\student}{\ensuremath{M}_{student}}
\newcommand{\baseline}{\ensuremath{M}_{baseline}}
\newcommand{\studenthard}{\ensuremath{M}_{student\_hard}}

\title{Memorization Dynamics in Knowledge Distillation for Language Models}

\author[1,4,*]{Jaydeep Borkar}
\author[2, *]{Karan Chadha}
\author[3,5,*]{Niloofar Mireshghallah}
\author[1,*]{Yuchen Zhang}
\author[1]{Irina-Elena Veliche}
\author[1]{Archi Mitra}
\author[4]{David A. Smith}
\author[1]{Zheng Xu}
\author[1]{Diego Garcia-Olano}

\affiliation[1]{Meta Superintelligence Labs}
\affiliation[2]{Meta Central Applied Science}
\affiliation[3]{FAIR at Meta, \\}
\addtolist[4]{Northeastern University}{\affiliationlist}{\affiliationformat}{}
\affiliation[5]{Carnegie Mellon University}

\contribution[*]{Work done at Meta}

\abstract{
Knowledge Distillation (KD) is increasingly adopted to transfer capabilities from large language models to smaller ones, offering significant improvements in efficiency and utility while often surpassing standard fine-tuning. Beyond performance, KD is also explored as a privacy-preserving mechanism to mitigate the risk of training data leakage. While training data memorization has been extensively studied in standard pre-training and fine-tuning settings, its dynamics in a knowledge distillation setup remain poorly understood. In this work, we study memorization across the KD pipeline using three large language model (LLM) families (Pythia, OLMo-2, Qwen-3) and three datasets (FineWeb, Wikitext, Nemotron-CC-v2). We find: (1) distilled models memorize significantly less training data than standard fine-tuning (reducing memorization by more than 50\%); (2) some examples are inherently easier to memorize and account for a large fraction of memorization during distillation (over ~95\%); (3) student memorization is predictable prior to distillation using features based on zlib entropy, KL divergence, and perplexity; and (4) while soft and hard distillation have similar overall memorization rates, hard distillation poses a greater risk: it inherits $2.7\times$ more teacher-specific examples than soft distillation. Overall, we demonstrate that distillation can provide both improved generalization and reduced memorization risks compared to standard fine-tuning.}

\date{\today}
\correspondence{\email{jaijborkar@gmail.com} \& \email{diegoolano@meta.com}}

\usepackage{float}
\usepackage{amssymb}

\begin{document}

\maketitle

\begin{figure*}[h]
    \centering
    \includegraphics[width=0.65\textwidth]{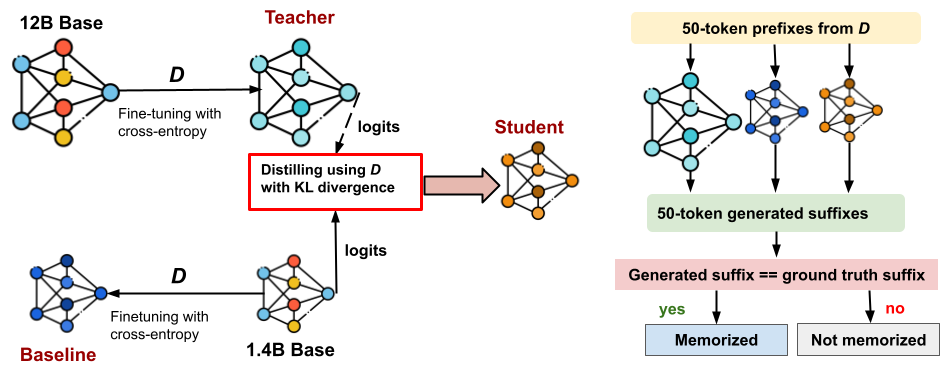}
    \caption{\textbf{Experimental framework.} \textbf{(Left)} Training setup: The Teacher and Baseline models are independently fine-tuned from Pythia 12B and Pythia 1.4B base models, respectively, on dataset $\dataset$ using cross-entropy. The Student is initialized from the Pythia 1.4B base model and distilled on the same dataset $\dataset$ to match the Teacher's logit distribution by minimizing KL divergence. \textbf{(Right)} Discoverable memorization evaluation: We prompt models with 50-token prefixes from training examples. An example is classified as \textit{memorized} if the model's greedy generation of the subsequent 50 tokens exactly matches the ground truth suffix.} 
    \label{fig:exp_setup}
\end{figure*}

\section{Introduction}
\label{section:intro}

Knowledge distillation (KD)~\citep{hinton2015distillingknowledgeneuralnetwork}, which transfers knowledge from larger teacher models to smaller student models, has been widely adopted in large language model (LLM) training due to its benefits in utility, efficiency and privacy~\citep{agarwal2023onpolicy, kodistillm}. KD is used to improve model quality by distilling knowledge from more powerful models, such as the DeepSeek-R1 distilled series~\citep{deepseekai2025deepseekr1incentivizingreasoningcapability} and the Gemma distilled models~\citep{team2024gemma}. Since training language models with billions of parameters from scratch is computationally expensive~\citep{compute}, KD has emerged as an effective approach to obtain higher-quality models with comparable or even reduced compute.

Beyond utility and efficiency, distillation is frequently cited as a potential defense against privacy vulnerabilities, specifically to protect the teacher's training data against privacy attacks~\citep{pate, Shejwalkar2021MembershipPF, 280000}. One such class of attacks is data extraction attacks, where an attacker can extract some portions of the training data from LLMs~\citep{carlini2019secretsharerevaluatingtesting, Carlini2020ExtractingTD, nasr2023scalableextractiontrainingdata}. While memorization and training data extraction have been extensively studied in standard pre-training~\citep{carlini2023quantifyingmemorizationneurallanguage, biderman2023emergentpredictablememorizationlarge, prashanth2025recitereconstructrecollectmemorization, morris2025languagemodelsmemorize, wei2025hubblemodelsuiteadvance} and fine-tuning~\citep{mireshghallah-etal-2022-empirical, borkar2023learndataleakageunlearning, zeng-etal-2024-exploring, borkar2025privacyrippleeffectsadding}, the mechanics of memorization remain poorly understood in a traditional KD setup. In such settings, the student is trained to mimic the teacher's distribution using the Kullback–Leibler (KL) divergence loss~\citep{Kullback1951OnIA}.

To address this gap, we systematically study training data memorization during knowledge distillation in a fine-tuning setup by examining both the amount and the characteristics of example sequences extracted from distilled models compared to standard fine-tuned baselines (trained with conventional cross-entropy loss). We distill models from the Pythia~\citep{pythia}, OLMo-2~\citep{olmo20252olmo2furious}, and Qwen-3~\citep{qwen3technicalreport} families using three distinct datasets, FineWeb~\citep{penedo2024finewebdatasetsdecantingweb}, WikiText~\citep{wikitext}, and the synthetic Nemotron-CC-v2 dataset~\citep{nvidia2025nvidianemotronnano2}. We report the following four main findings:
\begin{enumerate}
    \item Logit-level distillation using KL divergence significantly reduces memorization of training data and yields better generalization compared to standard fine-tuning. Crucially, the student recovers 78\% of the teacher’s generalization over standard fine-tuning while inheriting only 2\% of its memorization (Section~\ref{sec:memorize_less}).
    
    \item Certain examples are consistently memorized across models of varying sizes within the same family because they are \emph{inherently easier to memorize}. Furthermore, we find that distilled models preferentially memorize these \emph{easy-to-memorize} sequences (Section~\ref{subsec:easy_to_memorize}). 
    
    \item Memorization in student models is associated with features computable before distillation, and is predictable from them (Section~\ref{sec:pre_identifying}). Furthermore, we investigate the mechanism behind distillation's regularizing effect by analyzing sequence-level Shannon entropy and log probability (Figure~\ref{fig:entropy_prob} \& Section~\ref{section:why_distillation_reduces_mem}). 
    
    \item Logit-level (soft) and sequence-level (hard) distillation show significant memorization overlap, with the soft-distilled student capturing over 70\% of the examples memorized by the hard-distilled student. Despite this similarity, hard distillation poses a greater risk of inheriting memorization of \emph{difficult} examples from the Teacher (\textbf{2.7$\times$} more compared to soft distillation) (Section~\ref{sec:soft_vs_hard}).  
    
\end{enumerate}

\section{Background and Experimental Setup}
\label{sec:setup}

Our goal is to study memorization and extraction of training data within a knowledge distillation framework. We adopt the following definitions throughout the paper: (1) \textbf{Teacher model} ($\teacher$): the larger model that serves as a guide for training the smaller model; (2) \textbf{Student model} ($\student$): the smaller model trained to mimic the teacher; (3) \textbf{Baseline model} ($\baseline$): a model of the same size as the distilled student, but fine-tuned independently using standard cross-entropy loss; and (4) \textbf{Dataset} $\dataset$: the specific dataset used to train all three models. 

We now detail our experimental framework, comprising the knowledge distillation setup and our memorization evaluation metrics, as illustrated in Figure~\ref{fig:exp_setup}. Section~\ref{section:related_work} discusses prior work on memorization and privacy in knowledge distillation.

\paragraph{Knowledge Distillation (KD)}
Our main experiments use the Pythia models~\citep{pythia} and the FineWeb dataset~\citep{penedo2024finewebdatasetsdecantingweb}. To ensure the robustness of our findings, we extend our analysis to the OLMo-2~\citep{olmo20252olmo2furious} and Qwen-3~\citep{qwen3technicalreport} families, as well as the Nemotron-CC-v2~\citep{nvidia2025nvidianemotronnano2} and WikiText-103~\citep{wikitext} datasets. 

For the primary setup, we use $1\text{M}$ examples with a sequence length of 256 tokens from the July 2025 Common Crawl dump of FineWeb as our dataset $\dataset$. We first fine-tune the Pythia 12B base model on $\dataset$ using cross-entropy loss to obtain the teacher model $\teacher$. To obtain the student $\student$, we train the Pythia 1.4B base model on $\dataset$ using the teacher's guidance via forward KL divergence loss~\citep{Kullback1951OnIA, hinton2015distillingknowledgeneuralnetwork, kodistillm}, defined as:
\begin{equation}
\mathcal{L}_{\mathrm{KD}}
= T^2 \sum_{i=1}^{|\mathcal{V}|}
P_{\text{teacher}}^{\tau}(i)
\log \frac{P_{\text{teacher}}^{\tau}(i)}{P_{\text{student}}^{\tau}(i)},
\label{eqn:eqn1}
\end{equation}

where $T$ is the temperature parameter (set to $2.0$ in our experiments), and
\begin{equation}
P^{\tau}(i)
= \mathrm{softmax}\!\left(\frac{z_i}{T}\right)
= \frac{\exp(z_i/T)}{\sum_{j=1}^{|\mathcal{V}|} \exp(z_j/T)}
\label{eqn:eqn2}
\end{equation}
denotes the temperature-scaled probability of vocabulary token $i$, where $z_i$ is the pre-softmax logit corresponding to vocabulary token $i$. For comparison, we independently fine-tune the Pythia 1.4B base model on $\dataset$ using standard cross-entropy loss to obtain the baseline $\baseline$. All models are trained with a learning rate of $5 \times 10^{-5}$ and cosine decay. Both $\student$ and $\baseline$ are trained using comparable computational budgets. Validation loss and perplexity on a held-out set are reported in Table \ref{tab:loss}.

\paragraph{Memorization Evaluations}
We adopt the definition of \emph{discoverable} memorization from~\citet{nasr2023scalableextractiontrainingdata}. Formally, let $x \in \dataset$ be a training sequence split into a prefix $x_{1:k}$ and a suffix $x_{k+1:L}$, with $k=50$ and $L=100$. We define an example as memorized if the model's greedy generation exactly matches the ground truth suffix: $\mathcal{G}(x_{1:k}) = x_{k+1:L}$, where $\mathcal{G}(\cdot)$ represents the greedy decoding function of the model generating $L-k$ tokens. We evaluate this on 1M examples from our training dataset.

Section~\ref{sec:memorization_during_distillation} shows how much distilled models memorize compared to the teacher and the baseline, and provides insights into the characteristics of examples that are memorized by the student.
Section~\ref{sec:pre_identifying} shows that we can predict which examples the student will memorize prior to distillation. Section~\ref{section:why_distillation_reduces_mem} shows how distillation acts as a regularizer that reduces memorization.
Finally, Section~\ref{sec:soft_vs_hard} compares memorization in logit-level (soft) distillation with sequence-level (hard) distillation.

\begin{figure*}[t!]
    \centering
  \includegraphics[width=0.5\columnwidth]{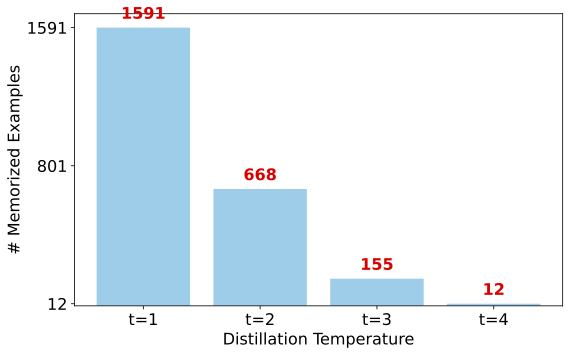}
  \caption{\textbf{Effect of Temperature on Memorization.} We find that increasing the temperature during distillation reduces memorization in the Student model.  
  }   
  \label{fig:temp}
\end{figure*}

\begin{table}[ht]
    \centering
    \caption{\textbf{Performance comparison across model families.} Validation loss and perplexity on the FineWeb dataset. Distilled students consistently outperform their respective baselines across Pythia, OLMo-2, and Qwen-3 families.}
    \label{tab:loss}
    
    \resizebox{0.63\columnwidth}{!}{ 
    \begin{tabular}{l cc c cc c cc}
        \toprule
         & \multicolumn{2}{c}{\textbf{Pythia}} & & \multicolumn{2}{c}{\textbf{OLMo-2}} & & \multicolumn{2}{c}{\textbf{Qwen-3}} \\
         \cmidrule{2-3} \cmidrule{5-6} \cmidrule{8-9}
        \textbf{Model} & \textbf{Loss} & \textbf{PPL} & & \textbf{Loss} & \textbf{PPL} & & \textbf{Loss} & \textbf{PPL} \\
        \midrule
        Teacher  & 2.75 & 15.66 & & 3.41 & 26.34 & & 3.34 & 23.49 \\
        Baseline & 2.87 & 17.69 & & 3.67 & 34.61 & & 3.65 & 33.23 \\
        Student  & \textbf{2.85} & \textbf{17.31} & & \textbf{3.44} & \textbf{28.15} & & \textbf{3.40} & \textbf{25.65} \\
        \bottomrule
    \end{tabular}
    }
\end{table}

\section{Memorization During Distillation}
\label{sec:memorization_during_distillation}
Contemporary knowledge distillation typically involves compressing massive models into smaller, efficient variants. For instance, the DeepSeek-R1 671B model serves as a  teacher to distill models as small as 1.5B~\citep{deepseekai2025deepseekr1incentivizingreasoningcapability}. Since memorization capacity correlates strongly with model scale~\citep{carlini2023quantifyingmemorizationneurallanguage}, these large  teachers inevitably memorize substantial portions of the training data. This raises two critical questions: First, to what extent does the student inherit the  teacher's generalization capabilities versus its memorization? Second, does the student memorize more than a  baseline model of the same size trained independently? To answer these, we analyze memorization across: (1) the distilled student model, (2) the independently fine-tuned  baseline, and (3) the  teacher model. This comparative analysis allows us to quantify the \emph{extent} of memorization in each model and characterize the \emph{specific properties} of the sequences they memorize. 

\subsection{Distilled Models Generalize Better and Memorize Less}
\label{sec:memorize_less}
We find that knowledge distillation \textbf{simultaneously reduces memorization and improves generalization} compared to standard fine-tuning. 

\paragraph{Memorization Reduction.} Table~\ref{tab:mem_stats} (Left) shows memorization rates for the Pythia family across three distinct datasets. On natural data, the student memorizes significantly less than the  baseline, reducing the rate by approximately 2.4$\times$ on FineWeb and 2.1$\times$ on Wikitext. While memorization rates are naturally lower on the synthetic Nemotron-CC-v2 dataset, the trend persists, with the student memorizing nearly an order of magnitude less than the  baseline (0.0012\% vs. 0.0091\%). As shown in Table~\ref{tab:mem_stats} (Right), this phenomenon is robust across architectures as the student consistently memorizes significantly less than the  baseline for both the OLMo-2 and Qwen-3 families. We observe similar findings in a pre-training distillation setup, which we report in section~\ref{subsec:pretrain_distillation}. We also find that increasing the distillation temperature $T$ reduces memorization in the student model (Figure~\ref{fig:temp}). These extraction results differ from \citet{students_parrot}, who report that lower temperatures are less vulnerable to membership inference attacks~\citep{lira}. We also observe the same reduction under looser notions of memorization, where the distilled student memorizes less than the fine-tuning baseline both for approximate memorization~\citep{ippolito-etal-2023-preventing} and for a longer 100-token prefix and suffix (Appendix~\ref{subsec:mem_definition_robustness}).

\paragraph{Better Generalization.} Crucially, this reduction in memorization does not come at the cost of model utility. As shown in Table~\ref{tab:loss}, the Pythia and OLMo-2 students achieve a lower validation loss and better perplexity compared to the  baseline. Section~\ref{section:extended_generalization_results} reports values for Pythia models trained on Wikitext. This suggests that distillation encourages the model to learn generalizable patterns from the  teacher rather than overfitting to specific training examples.

\paragraph{Inheriting  teacher's Generalization with Less Memorization.} Finally, we examine whether the student inherits the  teacher's specific memorization. We define \emph{memorization inheritance} as examples that are memorized by the  teacher and the student, but \textbf{not} by the  baseline. 
Prior work by~\citet{dankers-raunak-2025-memorization} on sequence distillation for machine translation found that students typically inherit their  teacher's memorization. In contrast, we find 1,955 examples memorized by  teacher but not by  baseline, and the student inherited only 18 (about 0.9\%) of them (Figure~\ref{fig:venn}). This confirms that the student learns the  teacher's general capabilities (as shown by better validation loss and perplexity than the  baseline in Table~\ref{tab:loss}) but successfully rejects the majority of the specific examples the  teacher exclusively memorized.

\begin{table*}[t]
    \centering
    \caption{\textbf{Memorization analysis.} \textbf{(Left)} Memorization rates of the Pythia family across natural (FineWeb, Wikitext) and synthetic (Nemotron-CC-v2) datasets. On FineWeb, the Pythia student memorizes $0.0706 \pm 0.0052\%$, compared to $0.1698 \pm 0.0081\%$ for the baseline and $0.3327 \pm 0.0113\%$ for the teacher, where each margin is the 95\% confidence-interval half-width. \textbf{(Right)} Memorization rates on the FineWeb dataset across different model families (Pythia, OLMo-2, Qwen-3). The student consistently memorizes less than the  baseline across all datasets and architectures. All  teacher models are trained for three epochs. student and  baseline models are trained for four epochs (Pythia) or five epochs (OLMo-2, Qwen-3).}
    \label{tab:mem_stats}
    
    \small
    
    \begin{minipage}[t]{0.48\textwidth}
        \centering
        
        \begin{tabular*}{\linewidth}{l @{\extracolsep{\fill}} lll}
            \toprule
            & \multicolumn{3}{c}{\textbf{Pythia Memorization Rate (\%)}} \\
            \cmidrule(lr){2-4}
            \textbf{Model} & \textbf{FineWeb} & \textbf{Wikitext} & \textbf{Nemotron} \\
            \midrule
             teacher (12B)   & 0.33 & 1.75 & 0.05 \\
            \midrule
             baseline (1.4B) & 0.17 & 0.21 & 0.0091 \\
            student (1.4B)  & \textbf{0.07} & \textbf{0.10} & \textbf{0.0012} \\
            \bottomrule
        \end{tabular*}
    \end{minipage}
    \hfill % Adds space between tables
    % --- RIGHT TABLE ---
    \begin{minipage}[t]{0.48\textwidth}
        \centering
        \begin{tabular*}{\linewidth}{l @{\extracolsep{\fill}} lll}
            \toprule
            & \multicolumn{3}{c}{\textbf{FineWeb Memorization Rate (\%)}} \\
            \cmidrule(lr){2-4}
            \textbf{Family} & \textbf{$\teacher$} & \textbf{$\baseline$} & \textbf{$\student$} \\ 
            \midrule
            Pythia  & 0.33 & 0.17 & \textbf{0.07} \\
            OLMo-2  & 8.90 & 0.40 & \textbf{0.09} \\
            Qwen-3  & 3.45 & 0.86 & \textbf{0.26} \\
            \bottomrule
        \end{tabular*}
    \end{minipage}
\end{table*}

\subsection{Some Examples are Easier to Memorize}
\label{subsec:easy_to_memorize}

\begin{figure}[H]
    \centering
    \begin{minipage}[b]{0.475\textwidth}
        \centering
        \includegraphics[width=0.8\linewidth]{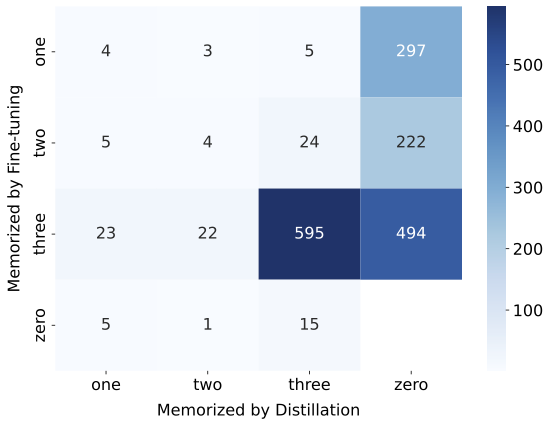}
        \caption{\textbf{Consistency heatmap.} We compare the consistency of memorization by the count from three independent runs for the  baseline (rows) and student (columns). The cell values represent the number of examples. The strong density in the (\textit{three}, \textit{three}) cell confirms that naturally "easy" examples are consistently memorized by both models. Conversely, the high count in (\textit{three}, \textit{zero}) highlights 494 examples that are consistently memorized by the  baseline but successfully suppressed (memorized in zero runs) by the student.}
        \label{fig:overlap_heatmap}
    \end{minipage}
    \hfill 
    \begin{minipage}[b]{0.475\textwidth}
        \centering
        \includegraphics[width=0.7\linewidth]{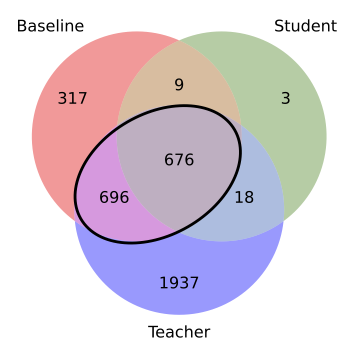}
        \caption{\textbf{Overlap of memorized examples.} The majority (80\%) of examples memorized by Pythia 1.4B  baseline are also memorized by the Pythia 12B  teacher. We term these consistently memorized examples as \emph{easy-to-memorize} (enclosed by a bold outline). The distilled student primarily memorizes a subset of these specific examples. To account for variance in training dynamics, we train three student and three  baseline models with different random seeds and report the union of memorized examples across these runs.}
        \label{fig:venn}
    \end{minipage}
\end{figure}

Memorization appears largely deterministic rather than stochastic, with specific \emph{easy} examples persisting across model scales and random seeds\footnote{For each seed, we vary the data order and model initialization while keeping all hyperparameters fixed.}. We observe a clear hierarchy where larger models retain the memorization of smaller ones: 96\% of examples memorized by Pythia 1B persist in the 1.4B  baseline, and the 12B  teacher captures approximately 80\% of the 1.4B  baseline's memorized set (as highlighted by the bold outline in Figure~\ref{fig:venn}). This determinism extends to initialization, where the  baseline consistently memorizes a core set of identical examples across all three independent training runs (row \emph{three} in Figure~\ref{fig:overlap_heatmap}). We term these consistently memorized examples \emph{easy-to-memorize}.

\paragraph{Why Are Some Examples Easier to Memorize?} Prior research has found training data duplication as a significant driver of memorization~\citep{lee-etal-2022-deduplicating, pmlr-v162-kandpal22a}. However, our training data sourced from the FineWeb doesn't contain any sequence-level duplicates~\citep{penedo2024finewebdatasetsdecantingweb}; therefore, duplication does not explain why some examples consistently get memorized over others in our setting. Instead, we investigate intrinsic properties of the text, specifically compressibility (measured via zlib entropy) and perplexity, which have been associated with memorization in prior studies~\citep{Carlini2020ExtractingTD, prashanth2025recitereconstructrecollectmemorization, borkar2025privacyrippleeffectsadding}. To compute zlib entropy, we first decode the tokenized sequence back into text and then measure the length (in bytes) of its zlib-compressed representation. Because decoding relies on a model-specific tokenizer, this metric is inherently model-dependent. 

\begin{figure*}[t]
    \centering
  \includegraphics[width=0.5\columnwidth]{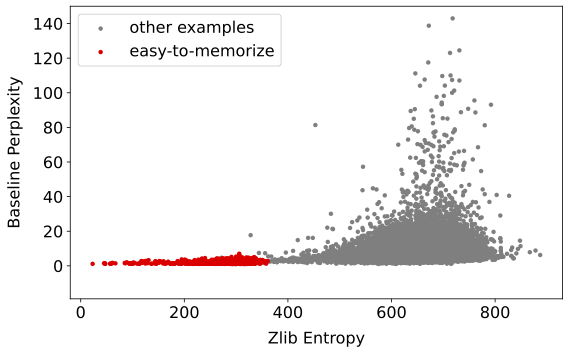}
  \caption{{\textbf{Intrinsic properties of \emph{easy-to-memorize} examples.} We plot zlib entropy versus  baseline perplexity for the \emph{easy-to-memorize} examples (red) compared to a random subset of 25,000 other examples (grey). They form a distinct cluster with significantly lower entropy and perplexity.}   }
  \label{fig:easy_to_mem}
\end{figure*}

We compute both metrics for examples from the \emph{easy-to-memorize} category and for a random subset of 25,000 other examples from our training dataset. As shown in Figure~\ref{fig:easy_to_mem}, we observe a distinct separation between the two groups. The \emph{easy-to-memorize} examples (red) form a tight cluster exhibiting significantly lower zlib entropy and  baseline perplexity compared to other training examples (grey). Section~\ref{subsec:appendix_why_some_examples_easy_memorize} reports similar findings on additional models and datasets.

Distillation acts as a strong regularizer within this hierarchy. The student model almost exclusively memorizes these \emph{easy} examples, as 95.7\% (676/706) of its memorization consists of examples shared by both the  teacher and  baseline (Figure~\ref{fig:venn}). However, distillation raises the bar for what gets memorized. Figure~\ref{fig:overlap_heatmap} shows 494 examples that the  baseline consistently memorized in all three runs, yet the student never memorized them. Similarly, the student fails to memorize 696 examples that are memorized both by the  baseline and  teacher (Figure~\ref{fig:venn}). Our findings generalize across datasets and model architectures (see Section~\ref{sec:extended_mem_overlap_analysis}). This suggests that distillation effectively removes a large portion of memorized data.

\subsubsection{Do All Models Memorize the Same \emph{Easy} Examples?}

From the previous section, we know that models from the same family memorize a consistent subset of examples. We now investigate whether these \emph{easy-to-memorize} examples are universal across different architectures. To test this, we trained OLMo-2 1B and Qwen-3 1.7B using the exact same data and hyperparameters as the Pythia 1.4B  baseline. While models within the same family show consistent memorization, we observe no overlap between Pythia, OLMo-2, and Qwen-3. In contrast, overlap persists among models within the OLMo-2 and Qwen-3 families (Figure~\ref{fig:overlap_extended}). This pattern persists at larger scales, with zero overlap observed between Pythia 12B, OLMo-2 7B, and Qwen-3 8B. This suggests that while \emph{easy} examples exist for any given model family, the specific set of examples selected for memorization is unique to the architecture's specific processing mechanisms (e.g., tokenization and attention biases).

\begin{figure}[H]
    \centering
    \begin{subfigure}[t]{0.32\columnwidth}
        \centering
        \includegraphics[width=0.8\linewidth]{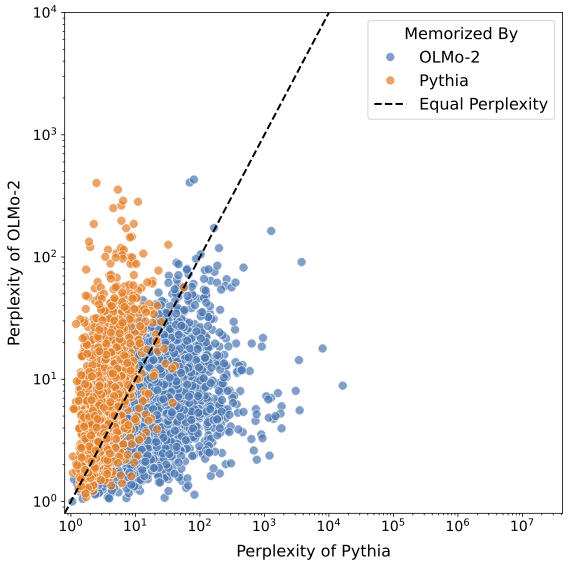}
        \caption{Pythia vs. OLMo-2}
        \label{fig:suba}
    \end{subfigure}
    \hfill
    \begin{subfigure}[t]{0.32\columnwidth}
        \centering
        \includegraphics[width=0.8\linewidth]{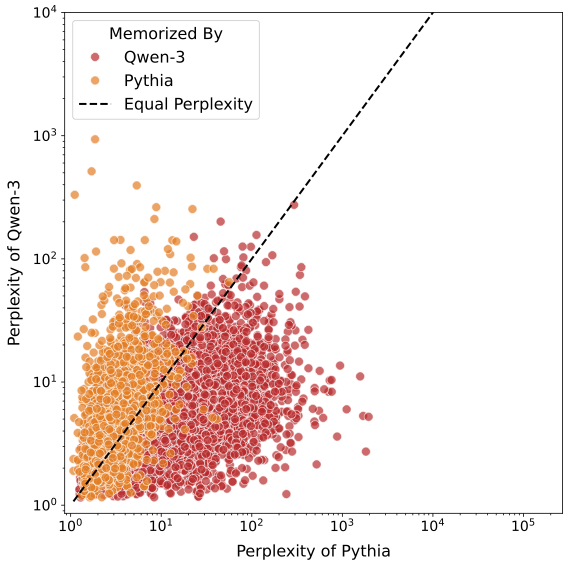}
        \caption{Pythia vs. Qwen-3}
        \label{fig:subb}
    \end{subfigure}
    \hfill
    \begin{subfigure}[t]{0.32\columnwidth}
        \centering
        \includegraphics[width=0.8\linewidth]{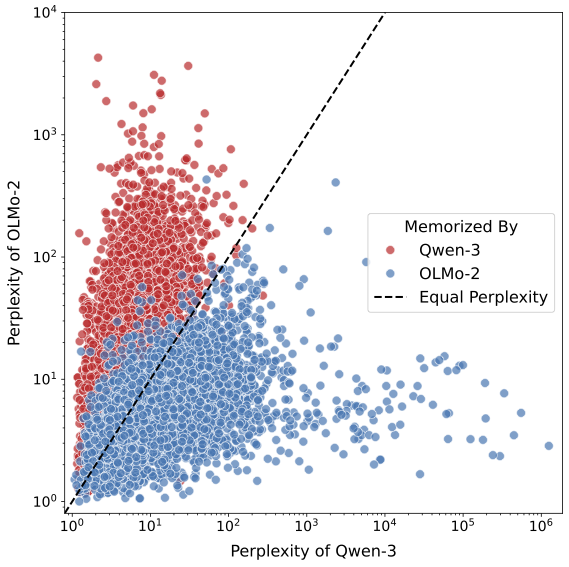}
        \caption{OLMo-2 vs. Qwen-3}
        \label{fig:subc}
    \end{subfigure}
    \caption{\textbf{Memorization preferences across model families.} We perform pairwise perplexity comparisons for examples memorized by Pythia, OLMo-2, and Qwen-3. In all three settings (a–c), we observe distinct, non-overlapping clusters. Examples memorized by one model are perceived as high perplexity by other architectures. This confirms that while all models target low-entropy examples, the specific selection of which examples to memorize is mutually exclusive and driven by model-specific inductive biases.}
    \label{fig:pythia_olmo_qwen}
\end{figure}

To explain this lack of overlap, we first determined whether the models perceive data complexity differently. We computed the zlib entropy (bits per token) for examples memorized by Pythia and compared this against the compression rates of Qwen-3 and OLMo-2 on the same text. We observed near-perfect alignment, with Pearson correlations of 0.95 (Pythia vs. Qwen-3), 0.96 (Pythia vs. OLMo-2), and 0.99 (Qwen-3 vs. OLMo-2). This confirms that all three architectures agree on which examples are \emph{easy} (low entropy) versus \emph{difficult}. However, despite fishing from the same pool of low-entropy examples, each model selects a different, non-overlapping subset to memorize. To understand this selection mechanism, we compared the perplexity of each model on examples memorized by the others. As shown in Figure~\ref{fig:pythia_olmo_qwen}, we observe distinct clusters: examples that one model finds trivial to memorize (low perplexity) are perceived as hard to memorize (high perplexity) by the other models, despite having low intrinsic entropy. This indicates that even though all models prefer memorizing simpler low entropy data, they don't agree on \emph{which} examples to memorize, which is largely driven by their inherent inductive biases.

\section{Identifying features associated with student memorization}
\label{sec:pre_identifying}
Current methods for detecting memorization typically require auditing a fully trained model. However, one may want to flag any potential memorized examples \emph{before} distillation begins so that these examples could be filtered out if needed. We therefore study features associated with student memorization that do not depend on the student model itself, and that could be leveraged to build a predictor to flag student memorization before training begins.

\subsection{Training a Memorization Classifier}

In a real-world distillation setup, it is common to have access to an already-trained large teacher model, which is then used to distill knowledge into smaller models of various sizes. It is sometimes also possible to have a baseline model of comparable size trained from scratch, often as part of a previous model release, but this is not always the case. We therefore study two sets of features: those that depend on a baseline model, and those that do not. We assume the
teacher and baseline models were trained on data similar to that used for distillation.

We identify 706 examples that are memorized by the student model after distillation; the remaining training examples are treated as non-memorized. We train a logistic regression classifier using memorized examples as the positive class and non-memorized examples as the negative class. We train two such classifiers: one using teacher perplexity, baseline perplexity, and the KL-divergence loss between the teacher and baseline as features, and one using only
teacher perplexity. Based on our findings from Section~\ref{subsec:easy_to_memorize}, we additionally include zlib entropy of the text, which has been associated with memorization in
prior studies, as a feature for both classifiers.

\begin{figure*}[t]
    \centering
    \includegraphics[width=0.98\textwidth]{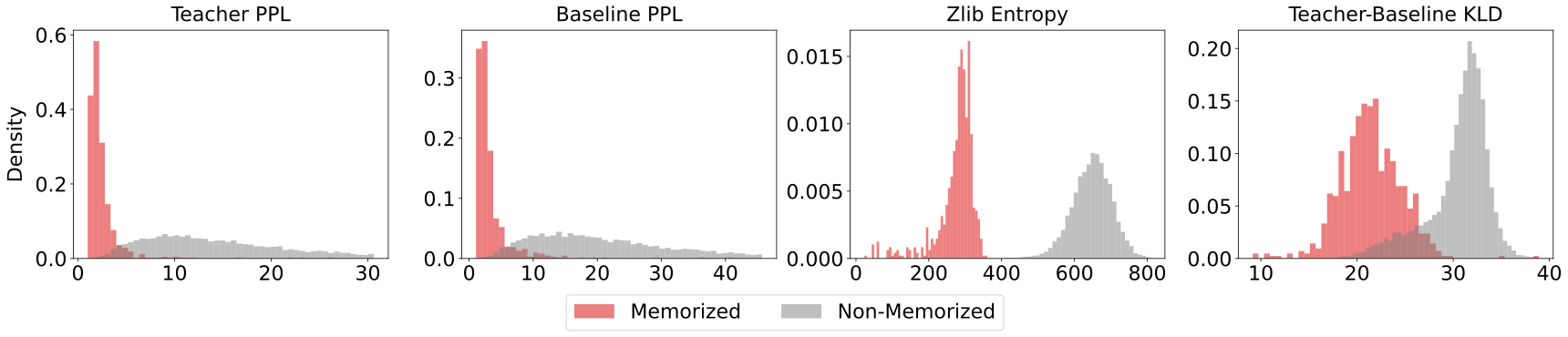}  
    \caption{\textbf{Feature distributions distinguishing memorized vs. non-memorized examples in student.} zlib entropy is the most significant predictor, followed by  baseline Perplexity, the KL divergence loss between the  teacher and  baseline, and  teacher perplexity. Across all metrics, lower values are consistently associated with memorized examples. Results are for Pythia models trained on the FineWeb.}
    \label{fig:all_metrics}
\end{figure*}

Our dataset has a 1:3 ratio of memorized to non-memorized examples. We reserve 30\% of the dataset for testing, and run 100 trials, resampling a new set of non-memorized examples in each trial. Our first classifier, using teacher perplexity, baseline perplexity, KL-divergence, and zlib entropy, achieves near-perfect discrimination between memorized and non-memorized examples, with an AUC-ROC of $0.9997 \pm 0.0005$. It further maintains a Recall of $1.0000 \pm 0.0000$, successfully identifying every memorized example in the test set across all 100 trials, with a negligible false positive rate (Precision: $0.9917 \pm 0.0059$). Our second classifier, using only
teacher perplexity and zlib entropy, achieves comparable, and in fact slightly better, performance (AUC $0.9998 \pm 0.0005$, Recall $1.0000 \pm 0.0000$, Precision $0.9940 \pm 0.0054$), despite
omitting baseline model features entirely.

Table~\ref{tab:feature_importance} reports feature weights for our first classifier. Zlib entropy is by far the most dominant feature (coefficient of $-4.50$), consistent with our findings from
Section~\ref{subsec:easy_to_memorize}, where lower zlib entropy is associated with \emph{easy-to-memorize} examples. Section~\ref{sec:extended_mem_classifier} reports the
performance of classifiers trained on these features across multiple datasets and model architectures. Figure~\ref{fig:all_metrics} shows the distributions of these features, with a clear separation between memorized and non-memorized examples; Section~\ref{sec:extended_all_features}
presents these distributions across additional datasets and model architectures.

\subsection{Removing memorized examples during distillation training}

Next, we study what happens if we remove the student memorized examples from the training dataset before starting the distillation training. Let $\dataset$ denote the training dataset used for distillation, and let $\hat{\dataset}_{mem}$ denote the set of examples memorized by the student. We remove these examples from the original training set to obtain $\dataset_{clean} = \dataset \setminus \hat{\dataset}_{mem}$. We use $\dataset_{clean}$ as our distillation dataset and  teacher $\teacher$ which is trained on $\dataset$ to get student $\student$. 

We find that this procedure results in the memorization of four new examples, consistent with observations in prior work~\citep{privacy_onion,borkar2025privacyrippleeffectsadding}. However, the overall number of memorized examples drops dramatically from 706 to four (a reduction of 99.4\%). This suggests that removing memorized examples can substantially reduce overall memorization during distillation.

\section{Why Does Distillation Reduce Memorization?}
\label{section:why_distillation_reduces_mem}
From Section~\ref{sec:memorize_less} we know that distilled student models memorize significantly less training data than  baseline models fine-tuned independently. We now investigate the mechanism behind this reduction. Specifically, we analyze the examples that the  baseline model memorizes but the student model doesn't.

\begin{figure}[t]
 \centering
  \includegraphics[width=0.8\columnwidth]{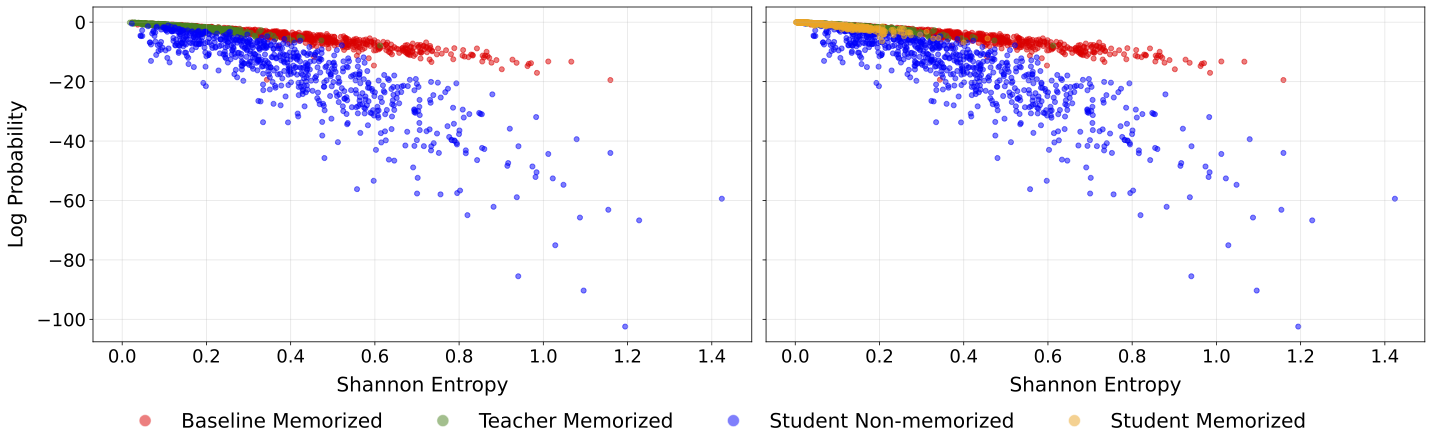}
  \caption{\textbf{Shannon Entropy vs. Log-Probability Analysis.} 
\textbf{Left:} The  teacher (Green) shows high probability (confidence) and low entropy (uncertainty) on its memorized examples. The  baseline (Red) forces high probability on high-entropy (uncertain) examples, resulting in \emph{forced memorization}. In contrast, the student (Blue) lowers its confidence on these uncertain examples, resulting in no memorization.
\textbf{Right:} The student's memorized examples (Orange) overlap closely with the  teacher (Green), confirming that the student memorizes only the \emph{easy} examples that it is very certain about.}
  \label{fig:entropy_prob}
\end{figure}

We identify 696 such examples where both the  baseline and  teacher (trained with cross-entropy) exhibit memorization, but the student (trained via KD) does not. We hypothesize that this could be due to the difference between the hard targets (one-hot labels) of cross-entropy and the soft targets (full probability distribution) of KL Divergence. To test this, we compute two metrics on the 50-token suffix of each example: (1) the sequence log-probability (confidence), and (2) the average Shannon entropy (intrinsic uncertainty) which can be written as

\[
H_t(x) = - \sum_{v \in \mathcal{V}} p_\theta(v \mid x_{<t}) \, \log p_\theta(v \mid x_{<t})
\]

\[
\bar{H}_\theta(x) = \frac{1}{K} \sum_{t=T-K}^{T-1} H_t(x)
\]

where $x = (x_1, \dots, x_T)$ is the tokenized sequence, $p_\theta(v \mid x_{<t})$ is the model probability for token $v$ at position $t$,  $\mathcal{V}$ is the vocabulary, and $K = 50$ is the number of last tokens used to compute the average. Figure~\ref{fig:entropy_prob} (Left) visualizes these metrics, showing three distinct behaviors.  

\textbf{Forced Memorization.} The  teacher model (Green) forms a tight cluster in the top-left region (high log-probability, low entropy). This indicates the 12B model is genuinely confident and finds these examples easy to memorize. In contrast, the  baseline model (Red) occupies a region of high log-probability but significantly higher entropy. This reveals a clear conflict: the 1.4B model lacks the capacity to model these complex examples naturally (shown by its high entropy), yet the cross-entropy loss forces it to assign high probability to the ground truth. This results in forced memorization, where the model overfits sequences it is uncertain about.

\textbf{Distillation as Regularization.} The student model (Blue) is limited by the same 1.4B capacity, so it remains uncertain (high entropy) about these difficult examples. However, unlike the  baseline, it exhibits low log-probability (i.e., no memorization). We attribute this to the training objective. While cross-entropy enforces a hard target, KL divergence allows the student to approximate the  teacher's distribution. When the student cannot match the  teacher's certainty on complex examples, the KD objective permits it to output a flatter, more uncertain distribution rather than forcing memorization.

Finally, we examine the examples the student \emph{does} memorize (Orange). These points cluster in the low-entropy region, overlapping heavily with the  teacher. This confirms that the student selectively memorizes only the examples that are simple enough to be learned with high confidence, effectively filtering out the high-entropy noise that the  baseline overfits.

\section{Soft vs. Hard Distillation}
\label{sec:soft_vs_hard}

Sequence-level knowledge distillation~\citep{kim-rush-2016-sequence} (or hard distillation) is an alternative approach where the student is trained on the  teacher’s generated output sequences using cross-entropy, rather than matching the  teacher's full probability distribution via KL divergence (soft distillation). This method is particularly relevant when the  teacher’s full output probabilities are inaccessible (e.g., black-box APIs). In this section, we investigate the memorization risks associated with this hard distillation method.   

\paragraph{Setup} We use the same  teacher $\teacher$ and dataset $\dataset$ as in our soft distillation experiments. For each example in $\dataset$, we prompt $\teacher$ with the first 50 tokens and generate 206 tokens using greedy decoding, resulting in a sequence of length $T=256$. This gives a synthetic dataset, $\dataset_{hard}$, used to train the student. We initialize the student parameters $\theta$ from the Pythia 1.4B base model and fine-tune on $\dataset_{hard}$ using cross-entropy loss:
\begin{equation}
    \mathcal{L}_{\text{hard}}(\theta) = -\mathbb{E}_{\hat{x} \sim \dataset_{hard}} \left[ \sum_{t=1}^{T} \log P_{\theta}(\hat{x}_t \mid \hat{x}_{<t}) \right]
\end{equation}
where $\hat{x}_t$ denotes the token at step $t$ and $\hat{x}_{<t}$ represents the preceding context.

We train $\studenthard$ using the same compute budget and learning rate as $\baseline$. To ensure a fair comparison, we evaluate performance on the LAMBADA~\citep{paperno-etal-2016-lambada} and Winogrande~\citep{winogrande} benchmarks using ~\citet{eval-harness} instead of measuring perplexity on a held-out set from the same distribution. Perplexity is misleading in this context because of the difference in training data: the  baseline optimizes on real data, while the student optimizes on the  teacher's synthetic outputs. As a result, the  baseline will naturally achieve lower perplexity on real validation sets, even if the student is more capable. Therefore, we use downstream tasks to measure actual utility.  

We observe that $\studenthard$ outperforms $\baseline$. Specifically, $\baseline$ achieves a perplexity of 9.41 and an accuracy of 51.85\% on LAMBADA, and an accuracy of 55.72\% on Winogrande. $\studenthard$ achieves a perplexity of 6.43 and an accuracy of 56.65\% on LAMBADA, along with an accuracy of 57.46\% on Winogrande. 
 
\begin{figure}[t]
 \centering
  \includegraphics[width=0.5\columnwidth]{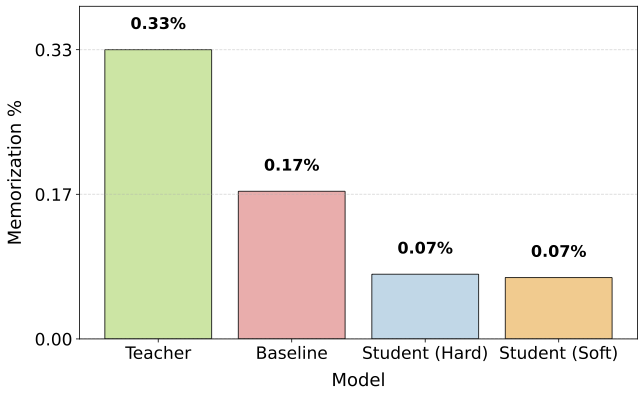}
  \caption{Hard-distilled and soft-distilled student models exhibit similar memorization rates, both significantly lower than the  baseline.}  
  \label{fig:hard_distill_stats}
\end{figure}

\begin{figure}[H]  
    \centering
    
    \begin{minipage}[t]{0.52\textwidth}
        \centering
        \includegraphics[width=0.8\linewidth, keepaspectratio]{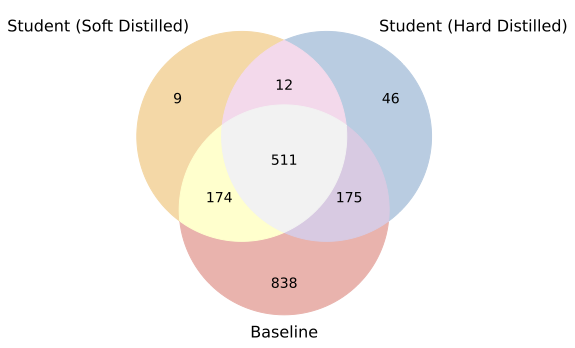}
        \caption{\textbf{Overlap between distillations.} Soft-distilled and Hard-distilled students memorize a lot of similar examples with a 70\% overlap.}
        \label{fig:venn_hard_soft}
    \end{minipage}
    \hfill  
    \begin{minipage}[t]{0.43\textwidth}
        \centering
        \includegraphics[width=0.8\linewidth, keepaspectratio]{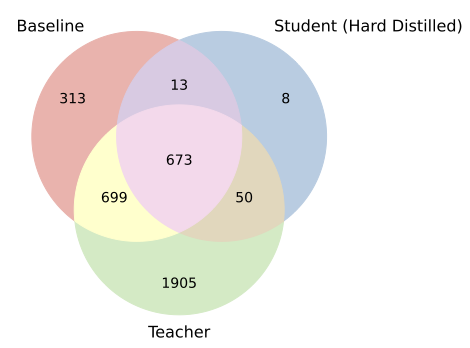}
        \caption{\textbf{Hard distillation dynamics.} Similar to soft-distillation, the majority of the examples (673) memorized by the hard-distilled student are \emph{easy-to-memorize} (i.e., they are memorized both by the  teacher and  baseline).}
        \label{fig:venn_hard_distilled}
    \end{minipage}
    
\end{figure}

\paragraph{Memorization Analysis.} As shown in Figure~\ref{fig:hard_distill_stats}, the hard-distilled student exhibits a memorization rate of 0.07\%, identical to the soft-distilled student and significantly lower than the  baseline model (0.17\%). Despite the different training objectives, there is a substantial overlap in what they memorize: approximately 70\% of the examples memorized by the hard-distilled student are also memorized by the soft-distilled student (Figure~\ref{fig:venn_hard_soft}). Furthermore, similar to the soft-distilled setup, the hard-distilled student primarily memorizes \emph{easy to memorize} examples (i.e., 90\% of these examples are also memorized by the  teacher and  baseline) (Figure~\ref{fig:venn_hard_distilled}).  

When we examine the examples memorized by one student but not the other, we find they are still largely dominated by the  baseline. For instance, the examples memorized by the soft student but missed by the hard student (and vice-versa) are frequently found in the  baseline set, further supporting that both students primarily target \emph{easy} examples.

\paragraph{Inheritance of Difficult Examples.} However, a distinct pattern emerges for the examples memorized exclusively by $\studenthard$. Figure~\ref{fig:venn_hard_soft} shows 46 examples memorized by $\studenthard$ that are not memorized by both $\student$ \& $\baseline$ (blue region). Our analysis reveals that the  teacher memorizes 80\% of this specific subset. We classify these as \emph{difficult} examples because they are not memorized by the  baseline but are inherited directly from the  teacher. This specific risk of hard distillation is quantified in Figure~\ref{fig:venn_hard_distilled} (brown region): the total amount of memorization inherited solely from the  teacher (memorized by  teacher and student but not  baseline) is 50 examples, an increase by a factor of 2.7 compared to the soft-distilled (Figure~\ref{fig:venn}). This suggests that while soft distillation effectively filters out these \emph{difficult} examples, hard distillation is more prone to memorizing them.

\section{Related Work}
\label{section:related_work}
\paragraph{Knowledge Distillation (KD)} was originally proposed to transfer knowledge from a large teacher to a smaller student by matching output distributions~\citep{hinton2015distillingknowledgeneuralnetwork, Bucila2006ModelC}. The standard objective minimizes the Kullback-Leibler (KL)~\citep{Kullback1951OnIA} divergence between the teacher's soft targets and the student's predictive distribution. While early methods focused on logit matching, subsequent approaches introduced intermediate feature matching~\citep{romero2015fitnetshintsdeepnets} to improve fidelity. In the context of LLMs, KD methods have evolved to address the challenges of autoregressive sequence generation. Sequence-level distillation~\citep{kim-rush-2016-sequence} methods use teacher-generated outputs instead of relying on the logits. More recently, specific techniques have been proposed to stabilize LLM distillation: MiniLLM~\citep{minillm} utilizes reverse KL divergence to prevent the student from over-approximating the teacher’s complex distribution, while Generalized Knowledge Distillation (GKD)~\citep{agarwal2024onpolicydistillationlanguagemodels} explores on-policy distillation to mitigate the distribution shift between teacher and student. \cite{kodistillm, ko2025distillm2} propose specialized objectives to improve LLM distillation efficiency. Other approaches, such as Distilling Step-by-Step~\citep{hsieh-etal-2023-distilling}, focus on transferring reasoning capabilities by distilling Chain-of-Thought rationales alongside final labels.

\paragraph{Memorization} \citet{carlini2019secretsharerevaluatingtesting} were the first to study the phenomenon of \emph{unintended} memorization in generative models. This led to numerous in-depth studies on memorization where pretrained language models regurgitate portions of their training data~\citep{Carlini2020ExtractingTD, tirumala, carlini2023quantifyingmemorizationneurallanguage,  nasr2023scalableextractiontrainingdata, biderman2023emergentpredictablememorizationlarge, huang2024demystifyingverbatimmemorizationlarge, prashanth2025recitereconstructrecollectmemorization, hayes-etal-2025-measuring, morris2025languagemodelsmemorize, ahmed2026extractingbooksproductionlanguage}, including studies on post-trained~\citep{barbero2025extractingalignmentdataopen} and fine-tuned models~\citep{mireshghallah-etal-2022-empirical, zeng-etal-2024-exploring, borkar2025privacyrippleeffectsadding}. However, despite these efforts on demystifying memorization, it remains poorly understood in a KD setup.

\paragraph{Memorization \& Privacy in Knowledge Distillation (KD)} Prior research on studying privacy during knowledge distillation includes membership inference attacks~\citep{students_parrot, zhang-etal-2025-membership, cui2025membershipinferenceattacksknowledge}, where the student can reveal information about the teacher's training data. \citet{distillationrobustifiesunlearning} find that distillation discards undesirable behavior (e.g., vulnerability to adversarial elicitation) while keeping the desired behavior intact. Most related to our work are the studies on data extraction in a KD setup: mainly \citet{dandekar-etal-2024-translation} who show that student machine translation models inherit a lot of the teacher's memorization during sequence-level distillation, \citet{zhang-etal-2025-membership} also study memorization inheritance, and their results indicate that students struggle to reproduce the teacher's entire memorization of 32 tokens, and \citet{singh-2025-teacher} similarly find that distillation can help reduce memorization. 

Unlike prior studies that focus on privacy leakage using membership inference attacks or memorization inheritance, our work systematically tracks memorization across the entire KD pipeline, including teacher, baseline, and student models. We measure both the amount and type of examples each model memorizes and show that student models primarily memorize \emph{easy-to-memorize examples}. Furthermore, we provide a mechanistic explanation for why KD acts as a regularizer to reduce memorization compared to standard fine-tuning.

\section{Conclusion}
We demonstrate that knowledge distillation improves model utility while significantly reducing training data memorization compared to standard fine-tuning. Our results show that distilled models primarily memorize \emph{easy-to-memorize} examples. Leveraging this finding, we show that high-risk examples can be predicted and removed prior to distillation, which substantially lowers memorization rates in the Student. We further explain why logit-level KD reduces memorization by analyzing log probabilities and Shannon entropy. Finally, we compare memorization risks between soft and hard distillation. We find that although both methods memorize largely the same examples, hard distillation poses a slightly higher risk due to greater memorization inheritance from the teacher.

\section{Acknowledgements}
We would like to thank Will Bullock, Dieuwke Hupkes, Saeed Mahloujifar, Shengyuan Hu, and colleagues from the Trust team for their feedback and discussions on this work. 

\clearpage
\newpage
\bibliographystyle{assets/plainnat}
\bibliography{custom, anthology}

\begin{thebibliography}{56}
\providecommand{\natexlab}[1]{#1}
\providecommand{\url}[1]{\texttt{#1}}
\expandafter\ifx\csname urlstyle\endcsname\relax
  \providecommand{\doi}[1]{doi: #1}\else
  \providecommand{\doi}{doi: \begingroup \urlstyle{rm}\Url}\fi

\bibitem[Agarwal et~al.(2023)Agarwal, Vieillard, Zhou, Stanczyk, Ramos, Geist, and Bachem]{agarwal2023onpolicy}
Rishabh Agarwal, Nino Vieillard, Yongchao Zhou, Piotr Stanczyk, Sabela Ramos, Matthieu Geist, and Olivier Bachem.
\newblock On-policy distillation of language models: Learning from self-generated mistakes, 2023.

\bibitem[Agarwal et~al.(2024)Agarwal, Vieillard, Zhou, Stanczyk, Ramos, Geist, and Bachem]{agarwal2024onpolicydistillationlanguagemodels}
Rishabh Agarwal, Nino Vieillard, Yongchao Zhou, Piotr Stanczyk, Sabela Ramos, Matthieu Geist, and Olivier Bachem.
\newblock On-policy distillation of language models: Learning from self-generated mistakes, 2024.
\newblock \url{https://arxiv.org/abs/2306.13649}.

\bibitem[Ahmed et~al.(2026)Ahmed, Cooper, Koyejo, and Liang]{ahmed2026extractingbooksproductionlanguage}
Ahmed Ahmed, A.~Feder Cooper, Sanmi Koyejo, and Percy Liang.
\newblock Extracting books from production language models, 2026.
\newblock \url{https://arxiv.org/abs/2601.02671}.

\bibitem[Barbero et~al.(2025)Barbero, Gu, Choquette-Choo, Sitawarin, Jagielski, Yona, Veličković, Shumailov, and Hayes]{barbero2025extractingalignmentdataopen}
Federico Barbero, Xiangming Gu, Christopher~A. Choquette-Choo, Chawin Sitawarin, Matthew Jagielski, Itay Yona, Petar Veličković, Ilia Shumailov, and Jamie Hayes.
\newblock Extracting alignment data in open models, 2025.
\newblock \url{https://arxiv.org/abs/2510.18554}.

\bibitem[Biderman et~al.(2023{\natexlab{a}})Biderman, Prashanth, Sutawika, Schoelkopf, Anthony, Purohit, and Raff]{biderman2023emergentpredictablememorizationlarge}
Stella Biderman, USVSN~Sai Prashanth, Lintang Sutawika, Hailey Schoelkopf, Quentin Anthony, Shivanshu Purohit, and Edward Raff.
\newblock Emergent and predictable memorization in large language models, 2023{\natexlab{a}}.
\newblock \url{https://arxiv.org/abs/2304.11158}.

\bibitem[Biderman et~al.(2023{\natexlab{b}})Biderman, Schoelkopf, Anthony, Bradley, O'Brien, Hallahan, Khan, Purohit, Prashanth, Raff, Skowron, Sutawika, and Van Der~Wal]{pythia}
Stella Biderman, Hailey Schoelkopf, Quentin Anthony, Herbie Bradley, Kyle O'Brien, Eric Hallahan, Mohammad~Aflah Khan, Shivanshu Purohit, USVSN~Sai Prashanth, Edward Raff, Aviya Skowron, Lintang Sutawika, and Oskar Van Der~Wal.
\newblock Pythia: a suite for analyzing large language models across training and scaling.
\newblock In \emph{Proceedings of the 40th International Conference on Machine Learning}, ICML'23. JMLR.org, 2023{\natexlab{b}}.

\bibitem[Borkar(2023)]{borkar2023learndataleakageunlearning}
Jaydeep Borkar.
\newblock What can we learn from data leakage and unlearning for law?, 2023.
\newblock \url{https://arxiv.org/abs/2307.10476}.

\bibitem[Borkar et~al.(2025)Borkar, Jagielski, Lee, Mireshghallah, Smith, and Choquette-Choo]{borkar2025privacyrippleeffectsadding}
Jaydeep Borkar, Matthew Jagielski, Katherine Lee, Niloofar Mireshghallah, David~A. Smith, and Christopher~A. Choquette-Choo.
\newblock Privacy ripple effects from adding or removing personal information in language model training, 2025.
\newblock \url{https://arxiv.org/abs/2502.15680}.

\bibitem[Bucila et~al.(2006)Bucila, Caruana, and Niculescu-Mizil]{Bucila2006ModelC}
Cristian Bucila, Rich Caruana, and Alexandru Niculescu-Mizil.
\newblock Model compression.
\newblock In \emph{Knowledge Discovery and Data Mining}, 2006.
\newblock \url{https://api.semanticscholar.org/CorpusID:11253972}.

\bibitem[Carlini et~al.(2019)Carlini, Liu, Úlfar Erlingsson, Kos, and Song]{carlini2019secretsharerevaluatingtesting}
Nicholas Carlini, Chang Liu, Úlfar Erlingsson, Jernej Kos, and Dawn Song.
\newblock The secret sharer: Evaluating and testing unintended memorization in neural networks, 2019.
\newblock \url{https://arxiv.org/abs/1802.08232}.

\bibitem[Carlini et~al.(2020)Carlini, Tram{\`e}r, Wallace, Jagielski, Herbert-Voss, Lee, Roberts, Brown, Song, Erlingsson, Oprea, and Raffel]{Carlini2020ExtractingTD}
Nicholas Carlini, Florian Tram{\`e}r, Eric Wallace, Matthew Jagielski, Ariel Herbert-Voss, Katherine Lee, Adam Roberts, Tom~B. Brown, Dawn~Xiaodong Song, {\'U}lfar Erlingsson, Alina Oprea, and Colin Raffel.
\newblock Extracting training data from large language models.
\newblock In \emph{USENIX Security Symposium}, 2020.
\newblock \url{https://api.semanticscholar.org/CorpusID:229156229}.

\bibitem[Carlini et~al.(2022{\natexlab{a}})Carlini, Chien, Nasr, Song, Terzis, and Tramèr]{lira}
Nicholas Carlini, Steve Chien, Milad Nasr, Shuang Song, Andreas Terzis, and Florian Tramèr.
\newblock Membership inference attacks from first principles.
\newblock In \emph{2022 IEEE Symposium on Security and Privacy (SP)}, pages 1897--1914, 2022{\natexlab{a}}.
\newblock \doi{10.1109/SP46214.2022.9833649}.

\bibitem[Carlini et~al.(2022{\natexlab{b}})Carlini, Terzis, Jagielski, Tramer, Papernot, and Zhang]{privacy_onion}
Nicholas Carlini, Andreas Terzis, Matthew Jagielski, Florian Tramer, Nicolas Papernot, and Chiyuan Zhang.
\newblock The privacy onion effect: memorization is relative.
\newblock In \emph{Proceedings of the 36th International Conference on Neural Information Processing Systems}, NIPS '22, Red Hook, NY, USA, 2022{\natexlab{b}}. Curran Associates Inc.
\newblock ISBN 9781713871088.

\bibitem[Carlini et~al.(2023)Carlini, Ippolito, Jagielski, Lee, Tramer, and Zhang]{carlini2023quantifyingmemorizationneurallanguage}
Nicholas Carlini, Daphne Ippolito, Matthew Jagielski, Katherine Lee, Florian Tramer, and Chiyuan Zhang.
\newblock Quantifying memorization across neural language models, 2023.
\newblock \url{https://arxiv.org/abs/2202.07646}.

\bibitem[Cui et~al.(2025)Cui, Zhang, and Pei]{cui2025membershipinferenceattacksknowledge}
Ziyao Cui, Minxing Zhang, and Jian Pei.
\newblock On membership inference attacks in knowledge distillation, 2025.
\newblock \url{https://arxiv.org/abs/2505.11837}.

\bibitem[Dandekar et~al.(2024)Dandekar, Xu, Xu, Ouyang, and Li]{dandekar-etal-2024-translation}
Chinmay Dandekar, Wenda Xu, Xi~Xu, Siqi Ouyang, and Lei Li.
\newblock Translation canvas: An explainable interface to pinpoint and analyze translation systems.
\newblock In Delia~Irazu Hernandez~Farias, Tom Hope, and Manling Li, editors, \emph{Proceedings of the 2024 Conference on Empirical Methods in Natural Language Processing: System Demonstrations}, pages 344--350, Miami, Florida, USA, November 2024. Association for Computational Linguistics.
\newblock \doi{10.18653/v1/2024.emnlp-demo.36}.
\newblock \url{https://aclanthology.org/2024.emnlp-demo.36/}.

\bibitem[Dankers and Raunak(2025)]{dankers-raunak-2025-memorization}
Verna Dankers and Vikas Raunak.
\newblock Memorization inheritance in sequence-level knowledge distillation for neural machine translation.
\newblock In Wanxiang Che, Joyce Nabende, Ekaterina Shutova, and Mohammad~Taher Pilehvar, editors, \emph{Proceedings of the 63rd Annual Meeting of the Association for Computational Linguistics (Volume 2: Short Papers)}, pages 760--774, Vienna, Austria, July 2025. Association for Computational Linguistics.
\newblock ISBN 979-8-89176-252-7.
\newblock \doi{10.18653/v1/2025.acl-short.61}.
\newblock \url{https://aclanthology.org/2025.acl-short.61/}.

\bibitem[et~al.(2025)]{nvidia2025nvidianemotronnano2}
NVIDIA et~al.
\newblock Nvidia nemotron nano 2: An accurate and efficient hybrid mamba-transformer reasoning model, 2025.
\newblock \url{https://arxiv.org/abs/2508.14444}.

\bibitem[Gao et~al.(2024)Gao, Tow, Abbasi, Biderman, Black, DiPofi, Foster, Golding, Hsu, Le~Noac'h, Li, McDonell, Muennighoff, Ociepa, Phang, Reynolds, Schoelkopf, Skowron, Sutawika, Tang, Thite, Wang, Wang, and Zou]{eval-harness}
Leo Gao, Jonathan Tow, Baber Abbasi, Stella Biderman, Sid Black, Anthony DiPofi, Charles Foster, Laurence Golding, Jeffrey Hsu, Alain Le~Noac'h, Haonan Li, Kyle McDonell, Niklas Muennighoff, Chris Ociepa, Jason Phang, Laria Reynolds, Hailey Schoelkopf, Aviya Skowron, Lintang Sutawika, Eric Tang, Anish Thite, Ben Wang, Kevin Wang, and Andy Zou.
\newblock The language model evaluation harness, 07 2024.
\newblock \url{https://zenodo.org/records/12608602}.

\bibitem[Gemma et~al.(2024)]{team2024gemma}
Team Gemma et~al.
\newblock Gemma 2: Improving open language models at a practical size, 2024.

\bibitem[Gu et~al.(2025)Gu, Dong, Wei, and Huang]{minillm}
Yuxian Gu, Li~Dong, Furu Wei, and Minlie Huang.
\newblock Minillm: Knowledge distillation of large language models, 2025.
\newblock \url{https://arxiv.org/abs/2306.08543}.

\bibitem[Guo et~al.(2025)]{deepseekai2025deepseekr1incentivizingreasoningcapability}
Daya Guo et~al.
\newblock Deepseek-r1: Incentivizing reasoning capability in llms via reinforcement learning, 2025.
\newblock \url{https://arxiv.org/abs/2501.12948}.

\bibitem[Hayes et~al.(2025)Hayes, Swanberg, Chaudhari, Yona, Shumailov, Nasr, Choquette-Choo, Lee, and Cooper]{hayes-etal-2025-measuring}
Jamie Hayes, Marika Swanberg, Harsh Chaudhari, Itay Yona, Ilia Shumailov, Milad Nasr, Christopher~A. Choquette-Choo, Katherine Lee, and A.~Feder Cooper.
\newblock Measuring memorization in language models via probabilistic extraction.
\newblock In Luis Chiruzzo, Alan Ritter, and Lu~Wang, editors, \emph{Proceedings of the 2025 Conference of the Nations of the Americas Chapter of the Association for Computational Linguistics: Human Language Technologies (Volume 1: Long Papers)}, pages 9266--9291, Albuquerque, New Mexico, April 2025. Association for Computational Linguistics.
\newblock ISBN 979-8-89176-189-6.
\newblock \doi{10.18653/v1/2025.naacl-long.469}.
\newblock \url{https://aclanthology.org/2025.naacl-long.469/}.

\bibitem[Hinton et~al.(2015)Hinton, Vinyals, and Dean]{hinton2015distillingknowledgeneuralnetwork}
Geoffrey Hinton, Oriol Vinyals, and Jeff Dean.
\newblock Distilling the knowledge in a neural network, 2015.
\newblock \url{https://arxiv.org/abs/1503.02531}.

\bibitem[Hoffmann et~al.(2022)Hoffmann, Borgeaud, Mensch, Buchatskaya, Cai, Rutherford, de~Las~Casas, Hendricks, Welbl, Clark, Hennigan, Noland, Millican, van~den Driessche, Damoc, Guy, Osindero, Simonyan, Elsen, Vinyals, Rae, and Sifre]{compute}
Jordan Hoffmann, Sebastian Borgeaud, Arthur Mensch, Elena Buchatskaya, Trevor Cai, Eliza Rutherford, Diego de~Las~Casas, Lisa~Anne Hendricks, Johannes Welbl, Aidan Clark, Tom Hennigan, Eric Noland, Katie Millican, George van~den Driessche, Bogdan Damoc, Aurelia Guy, Simon Osindero, Karen Simonyan, Erich Elsen, Oriol Vinyals, Jack~W. Rae, and Laurent Sifre.
\newblock Training compute-optimal large language models.
\newblock In \emph{Proceedings of the 36th International Conference on Neural Information Processing Systems}, NIPS '22, Red Hook, NY, USA, 2022. Curran Associates Inc.
\newblock ISBN 9781713871088.

\bibitem[Hsieh et~al.(2023)Hsieh, Li, Yeh, Nakhost, Fujii, Ratner, Krishna, Lee, and Pfister]{hsieh-etal-2023-distilling}
Cheng-Yu Hsieh, Chun-Liang Li, Chih-kuan Yeh, Hootan Nakhost, Yasuhisa Fujii, Alex Ratner, Ranjay Krishna, Chen-Yu Lee, and Tomas Pfister.
\newblock Distilling step-by-step! outperforming larger language models with less training data and smaller model sizes.
\newblock In Anna Rogers, Jordan Boyd-Graber, and Naoaki Okazaki, editors, \emph{Findings of the Association for Computational Linguistics: ACL 2023}, pages 8003--8017, Toronto, Canada, July 2023. Association for Computational Linguistics.
\newblock \doi{10.18653/v1/2023.findings-acl.507}.
\newblock \url{https://aclanthology.org/2023.findings-acl.507/}.

\bibitem[Huang et~al.(2024)Huang, Yang, and Potts]{huang2024demystifyingverbatimmemorizationlarge}
Jing Huang, Diyi Yang, and Christopher Potts.
\newblock Demystifying verbatim memorization in large language models, 2024.
\newblock \url{https://arxiv.org/abs/2407.17817}.

\bibitem[Ippolito et~al.(2023)Ippolito, Tramer, Nasr, Zhang, Jagielski, Lee, Choquette~Choo, and Carlini]{ippolito-etal-2023-preventing}
Daphne Ippolito, Florian Tramer, Milad Nasr, Chiyuan Zhang, Matthew Jagielski, Katherine Lee, Christopher Choquette~Choo, and Nicholas Carlini.
\newblock Preventing generation of verbatim memorization in language models gives a false sense of privacy.
\newblock In C.~Maria Keet, Hung-Yi Lee, and Sina Zarrie{\ss}, editors, \emph{Proceedings of the 16th International Natural Language Generation Conference}, pages 28--53, Prague, Czechia, September 2023. Association for Computational Linguistics.
\newblock \doi{10.18653/v1/2023.inlg-main.3}.
\newblock \url{https://aclanthology.org/2023.inlg-main.3/}.

\bibitem[Jagielski et~al.(2023)Jagielski, Nasr, Lee, Choquette-Choo, Carlini, and Tram\`{e}r]{students_parrot}
Matthew Jagielski, Milad Nasr, Katherine Lee, Christopher Choquette-Choo, Nicholas Carlini, and Florian Tram\`{e}r.
\newblock Students parrot their teachers: membership inference on model distillation.
\newblock In \emph{Proceedings of the 37th International Conference on Neural Information Processing Systems}, NIPS '23, Red Hook, NY, USA, 2023. Curran Associates Inc.

\bibitem[Kandpal et~al.(2022)Kandpal, Wallace, and Raffel]{pmlr-v162-kandpal22a}
Nikhil Kandpal, Eric Wallace, and Colin Raffel.
\newblock Deduplicating training data mitigates privacy risks in language models.
\newblock In Kamalika Chaudhuri, Stefanie Jegelka, Le~Song, Csaba Szepesvari, Gang Niu, and Sivan Sabato, editors, \emph{Proceedings of the 39th International Conference on Machine Learning}, volume 162 of \emph{Proceedings of Machine Learning Research}, pages 10697--10707. PMLR, 17--23 Jul 2022.
\newblock \url{https://proceedings.mlr.press/v162/kandpal22a.html}.

\bibitem[Kim and Rush(2016)]{kim-rush-2016-sequence}
Yoon Kim and Alexander~M. Rush.
\newblock Sequence-level knowledge distillation.
\newblock In Jian Su, Kevin Duh, and Xavier Carreras, editors, \emph{Proceedings of the 2016 Conference on Empirical Methods in Natural Language Processing}, pages 1317--1327, Austin, Texas, November 2016. Association for Computational Linguistics.
\newblock \doi{10.18653/v1/D16-1139}.
\newblock \url{https://aclanthology.org/D16-1139/}.

\bibitem[Ko et~al.()Ko, Kim, Chen, and Yun]{kodistillm}
Jongwoo Ko, Sungnyun Kim, Tianyi Chen, and Se-Young Yun.
\newblock Distillm: Towards streamlined distillation for large language models.
\newblock In \emph{Forty-first International Conference on Machine Learning}.

\bibitem[Ko et~al.(2025)Ko, Chen, Kim, Ding, Liang, Zharkov, and Yun]{ko2025distillm2}
Jongwoo Ko, Tianyi Chen, Sungnyun Kim, Tianyu Ding, Luming Liang, Ilya Zharkov, and Se-Young Yun.
\newblock Distillm-2: A contrastive approach boosts the distillation of llms.
\newblock \emph{arXiv preprint arXiv:2503.07067}, 2025.

\bibitem[Kullback and Leibler(1951)]{Kullback1951OnIA}
Solomon Kullback and R.~A. Leibler.
\newblock On information and sufficiency.
\newblock \emph{Annals of Mathematical Statistics}, 22:\penalty0 79--86, 1951.
\newblock \url{https://api.semanticscholar.org/CorpusID:120349231}.

\bibitem[Lee et~al.(2025)Lee, Foote, Infanger, Shor, Kamath, Goldman-Wetzler, Woodworth, Cloud, and Turner]{distillationrobustifiesunlearning}
Bruce~W. Lee, Addie Foote, Alex Infanger, Leni Shor, Harish Kamath, Jacob Goldman-Wetzler, Bryce Woodworth, Alex Cloud, and Alexander~Matt Turner.
\newblock Distillation robustifies unlearning, 2025.
\newblock \url{https://arxiv.org/abs/2506.06278}.

\bibitem[Lee et~al.(2022)Lee, Ippolito, Nystrom, Zhang, Eck, Callison-Burch, and Carlini]{lee-etal-2022-deduplicating}
Katherine Lee, Daphne Ippolito, Andrew Nystrom, Chiyuan Zhang, Douglas Eck, Chris Callison-Burch, and Nicholas Carlini.
\newblock Deduplicating training data makes language models better.
\newblock In Smaranda Muresan, Preslav Nakov, and Aline Villavicencio, editors, \emph{Proceedings of the 60th Annual Meeting of the Association for Computational Linguistics (Volume 1: Long Papers)}, pages 8424--8445, Dublin, Ireland, May 2022. Association for Computational Linguistics.
\newblock \doi{10.18653/v1/2022.acl-long.577}.
\newblock \url{https://aclanthology.org/2022.acl-long.577/}.

\bibitem[Merity et~al.(2016)Merity, Xiong, Bradbury, and Socher]{wikitext}
Stephen Merity, Caiming Xiong, James Bradbury, and Richard Socher.
\newblock Pointer sentinel mixture models, 2016.

\bibitem[Mireshghallah et~al.(2022)Mireshghallah, Uniyal, Wang, Evans, and Berg-Kirkpatrick]{mireshghallah-etal-2022-empirical}
Fatemehsadat Mireshghallah, Archit Uniyal, Tianhao Wang, David Evans, and Taylor Berg-Kirkpatrick.
\newblock An empirical analysis of memorization in fine-tuned autoregressive language models.
\newblock In Yoav Goldberg, Zornitsa Kozareva, and Yue Zhang, editors, \emph{Proceedings of the 2022 Conference on Empirical Methods in Natural Language Processing}, pages 1816--1826, Abu Dhabi, United Arab Emirates, December 2022. Association for Computational Linguistics.
\newblock \doi{10.18653/v1/2022.emnlp-main.119}.
\newblock \url{https://aclanthology.org/2022.emnlp-main.119/}.

\bibitem[Morris et~al.(2025)Morris, Sitawarin, Guo, Kokhlikyan, Suh, Rush, Chaudhuri, and Mahloujifar]{morris2025languagemodelsmemorize}
John~X. Morris, Chawin Sitawarin, Chuan Guo, Narine Kokhlikyan, G.~Edward Suh, Alexander~M. Rush, Kamalika Chaudhuri, and Saeed Mahloujifar.
\newblock How much do language models memorize?, 2025.
\newblock \url{https://arxiv.org/abs/2505.24832}.

\bibitem[Nasr et~al.(2023)Nasr, Carlini, Hayase, Jagielski, Cooper, Ippolito, Choquette-Choo, Wallace, Tramèr, and Lee]{nasr2023scalableextractiontrainingdata}
Milad Nasr, Nicholas Carlini, Jonathan Hayase, Matthew Jagielski, A.~Feder Cooper, Daphne Ippolito, Christopher~A. Choquette-Choo, Eric Wallace, Florian Tramèr, and Katherine Lee.
\newblock Scalable extraction of training data from (production) language models, 2023.
\newblock \url{https://arxiv.org/abs/2311.17035}.

\bibitem[OLMo et~al.(2025)OLMo, Walsh, Soldaini, Groeneveld, Lo, Arora, Bhagia, Gu, Huang, Jordan, Lambert, Schwenk, Tafjord, Anderson, Atkinson, Brahman, Clark, Dasigi, Dziri, Ettinger, Guerquin, Heineman, Ivison, Koh, Liu, Malik, Merrill, Miranda, Morrison, Murray, Nam, Poznanski, Pyatkin, Rangapur, Schmitz, Skjonsberg, Wadden, Wilhelm, Wilson, Zettlemoyer, Farhadi, Smith, and Hajishirzi]{olmo20252olmo2furious}
Team OLMo, Pete Walsh, Luca Soldaini, Dirk Groeneveld, Kyle Lo, Shane Arora, Akshita Bhagia, Yuling Gu, Shengyi Huang, Matt Jordan, Nathan Lambert, Dustin Schwenk, Oyvind Tafjord, Taira Anderson, David Atkinson, Faeze Brahman, Christopher Clark, Pradeep Dasigi, Nouha Dziri, Allyson Ettinger, Michal Guerquin, David Heineman, Hamish Ivison, Pang~Wei Koh, Jiacheng Liu, Saumya Malik, William Merrill, Lester James~V. Miranda, Jacob Morrison, Tyler Murray, Crystal Nam, Jake Poznanski, Valentina Pyatkin, Aman Rangapur, Michael Schmitz, Sam Skjonsberg, David Wadden, Christopher Wilhelm, Michael Wilson, Luke Zettlemoyer, Ali Farhadi, Noah~A. Smith, and Hannaneh Hajishirzi.
\newblock 2 olmo 2 furious, 2025.
\newblock \url{https://arxiv.org/abs/2501.00656}.

\bibitem[Paperno et~al.(2016)Paperno, Kruszewski, Lazaridou, Pham, Bernardi, Pezzelle, Baroni, Boleda, and Fern{\'a}ndez]{paperno-etal-2016-lambada}
Denis Paperno, Germ{\'a}n Kruszewski, Angeliki Lazaridou, Ngoc~Quan Pham, Raffaella Bernardi, Sandro Pezzelle, Marco Baroni, Gemma Boleda, and Raquel Fern{\'a}ndez.
\newblock The {LAMBADA} dataset: Word prediction requiring a broad discourse context.
\newblock In Katrin Erk and Noah~A. Smith, editors, \emph{Proceedings of the 54th Annual Meeting of the Association for Computational Linguistics (Volume 1: Long Papers)}, pages 1525--1534, Berlin, Germany, August 2016. Association for Computational Linguistics.
\newblock \doi{10.18653/v1/P16-1144}.
\newblock \url{https://aclanthology.org/P16-1144/}.

\bibitem[Papernot et~al.(2018)Papernot, Song, Mironov, Raghunathan, Talwar, and Úlfar Erlingsson]{pate}
Nicolas Papernot, Shuang Song, Ilya Mironov, Ananth Raghunathan, Kunal Talwar, and Úlfar Erlingsson.
\newblock Scalable private learning with pate, 2018.
\newblock \url{https://arxiv.org/abs/1802.08908}.

\bibitem[Penedo et~al.(2024)Penedo, Kydlíček, allal, Lozhkov, Mitchell, Raffel, Werra, and Wolf]{penedo2024finewebdatasetsdecantingweb}
Guilherme Penedo, Hynek Kydlíček, Loubna~Ben allal, Anton Lozhkov, Margaret Mitchell, Colin Raffel, Leandro~Von Werra, and Thomas Wolf.
\newblock The fineweb datasets: Decanting the web for the finest text data at scale, 2024.
\newblock \url{https://arxiv.org/abs/2406.17557}.

\bibitem[Prashanth et~al.(2025)Prashanth, Deng, O'Brien, V, Khan, Borkar, Choquette-Choo, Fuehne, Biderman, Ke, Lee, and Saphra]{prashanth2025recitereconstructrecollectmemorization}
USVSN~Sai Prashanth, Alvin Deng, Kyle O'Brien, Jyothir~S V, Mohammad~Aflah Khan, Jaydeep Borkar, Christopher~A. Choquette-Choo, Jacob~Ray Fuehne, Stella Biderman, Tracy Ke, Katherine Lee, and Naomi Saphra.
\newblock Recite, reconstruct, recollect: Memorization in lms as a multifaceted phenomenon, 2025.
\newblock \url{https://arxiv.org/abs/2406.17746}.

\bibitem[Romero et~al.(2015)Romero, Ballas, Kahou, Chassang, Gatta, and Bengio]{romero2015fitnetshintsdeepnets}
Adriana Romero, Nicolas Ballas, Samira~Ebrahimi Kahou, Antoine Chassang, Carlo Gatta, and Yoshua Bengio.
\newblock Fitnets: Hints for thin deep nets, 2015.
\newblock \url{https://arxiv.org/abs/1412.6550}.

\bibitem[Sakaguchi et~al.(2021)Sakaguchi, Bras, Bhagavatula, and Choi]{winogrande}
Keisuke Sakaguchi, Ronan~Le Bras, Chandra Bhagavatula, and Yejin Choi.
\newblock Winogrande: an adversarial winograd schema challenge at scale.
\newblock \emph{Commun. ACM}, 64\penalty0 (9):\penalty0 99–106, August 2021.
\newblock ISSN 0001-0782.
\newblock \doi{10.1145/3474381}.
\newblock \url{https://doi.org/10.1145/3474381}.

\bibitem[Shejwalkar and Houmansadr(2021)]{Shejwalkar2021MembershipPF}
Virat Shejwalkar and Amir Houmansadr.
\newblock Membership privacy for machine learning models through knowledge transfer.
\newblock In \emph{AAAI Conference on Artificial Intelligence}, 2021.
\newblock \url{https://api.semanticscholar.org/CorpusID:235349092}.

\bibitem[Shokri et~al.(2017)Shokri, Stronati, Song, and Shmatikov]{MIA_OG}
Reza Shokri, Marco Stronati, Congzheng Song, and Vitaly Shmatikov.
\newblock Membership inference attacks against machine learning models.
\newblock In \emph{2017 IEEE Symposium on Security and Privacy (SP)}, pages 3--18, 2017.
\newblock \doi{10.1109/SP.2017.41}.

\bibitem[Singh(2025)]{singh-2025-teacher}
Simardeep Singh.
\newblock From teacher to student: Tracking memorization through model distillation.
\newblock In Robin Jia, Eric Wallace, Yangsibo Huang, Tiago Pimentel, Pratyush Maini, Verna Dankers, Johnny Wei, and Pietro Lesci, editors, \emph{Proceedings of the First Workshop on Large Language Model Memorization (L2M2)}, pages 78--82, Vienna, Austria, August 2025. Association for Computational Linguistics.
\newblock ISBN 979-8-89176-278-7.
\newblock \doi{10.18653/v1/2025.l2m2-1.6}.
\newblock \url{https://aclanthology.org/2025.l2m2-1.6/}.

\bibitem[Tang et~al.(2022)Tang, Mahloujifar, Song, Shejwalkar, Nasr, Houmansadr, and Mittal]{280000}
Xinyu Tang, Saeed Mahloujifar, Liwei Song, Virat Shejwalkar, Milad Nasr, Amir Houmansadr, and Prateek Mittal.
\newblock Mitigating membership inference attacks by {Self-Distillation} through a novel ensemble architecture.
\newblock In \emph{31st USENIX Security Symposium (USENIX Security 22)}, pages 1433--1450, Boston, MA, August 2022. USENIX Association.
\newblock ISBN 978-1-939133-31-1.
\newblock \url{https://www.usenix.org/conference/usenixsecurity22/presentation/tang}.

\bibitem[Tirumala et~al.(2022)Tirumala, Markosyan, Zettlemoyer, and Aghajanyan]{tirumala}
Kushal Tirumala, Aram~H. Markosyan, Luke Zettlemoyer, and Armen Aghajanyan.
\newblock Memorization without overfitting: analyzing the training dynamics of large language models.
\newblock In \emph{Proceedings of the 36th International Conference on Neural Information Processing Systems}, NIPS '22, Red Hook, NY, USA, 2022. Curran Associates Inc.
\newblock ISBN 9781713871088.

\bibitem[Wei et~al.(2025)Wei, Godbole, Khan, Wang, Zhu, Flemings, Kashyap, Gummadi, Neiswanger, and Jia]{wei2025hubblemodelsuiteadvance}
Johnny Tian-Zheng Wei, Ameya Godbole, Mohammad~Aflah Khan, Ryan Wang, Xiaoyuan Zhu, James Flemings, Nitya Kashyap, Krishna~P. Gummadi, Willie Neiswanger, and Robin Jia.
\newblock Hubble: a model suite to advance the study of llm memorization, 2025.
\newblock \url{https://arxiv.org/abs/2510.19811}.

\bibitem[Yang et~al.(2025)Yang, Li, Yang, Zhang, Hui, Zheng, Yu, Gao, Huang, Lv, Zheng, Liu, Zhou, Huang, Hu, Ge, Wei, Lin, Tang, Yang, Tu, Zhang, Yang, Yang, Zhou, Zhou, Lin, Dang, Bao, Yang, Yu, Deng, Li, Xue, Li, Zhang, Wang, Zhu, Men, Gao, Liu, Luo, Li, Tang, Yin, Ren, Wang, Zhang, Ren, Fan, Su, Zhang, Zhang, Wan, Liu, Wang, Cui, Zhang, Zhou, and Qiu]{qwen3technicalreport}
An~Yang, Anfeng Li, Baosong Yang, Beichen Zhang, Binyuan Hui, Bo~Zheng, Bowen Yu, Chang Gao, Chengen Huang, Chenxu Lv, Chujie Zheng, Dayiheng Liu, Fan Zhou, Fei Huang, Feng Hu, Hao Ge, Haoran Wei, Huan Lin, Jialong Tang, Jian Yang, Jianhong Tu, Jianwei Zhang, Jianxin Yang, Jiaxi Yang, Jing Zhou, Jingren Zhou, Junyang Lin, Kai Dang, Keqin Bao, Kexin Yang, Le~Yu, Lianghao Deng, Mei Li, Mingfeng Xue, Mingze Li, Pei Zhang, Peng Wang, Qin Zhu, Rui Men, Ruize Gao, Shixuan Liu, Shuang Luo, Tianhao Li, Tianyi Tang, Wenbiao Yin, Xingzhang Ren, Xinyu Wang, Xinyu Zhang, Xuancheng Ren, Yang Fan, Yang Su, Yichang Zhang, Yinger Zhang, Yu~Wan, Yuqiong Liu, Zekun Wang, Zeyu Cui, Zhenru Zhang, Zhipeng Zhou, and Zihan Qiu.
\newblock Qwen3 technical report, 2025.
\newblock \url{https://arxiv.org/abs/2505.09388}.

\bibitem[Zeng et~al.(2024)Zeng, Li, Ren, Liu, Xu, He, Xing, Wang, Tang, and Yin]{zeng-etal-2024-exploring}
Shenglai Zeng, Yaxin Li, Jie Ren, Yiding Liu, Han Xu, Pengfei He, Yue Xing, Shuaiqiang Wang, Jiliang Tang, and Dawei Yin.
\newblock Exploring memorization in fine-tuned language models.
\newblock In Lun-Wei Ku, Andre Martins, and Vivek Srikumar, editors, \emph{Proceedings of the 62nd Annual Meeting of the Association for Computational Linguistics (Volume 1: Long Papers)}, pages 3917--3948, Bangkok, Thailand, August 2024. Association for Computational Linguistics.
\newblock \doi{10.18653/v1/2024.acl-long.216}.
\newblock \url{https://aclanthology.org/2024.acl-long.216/}.

\bibitem[Zhang et~al.(2025)Zhang, Shahin~Shamsabadi, Lu, Cai, and Haddadi]{zhang-etal-2025-membership}
Ziqi Zhang, Ali Shahin~Shamsabadi, Hanxiao Lu, Yifeng Cai, and Hamed Haddadi.
\newblock Membership and memorization in {LLM} knowledge distillation.
\newblock In Christos Christodoulopoulos, Tanmoy Chakraborty, Carolyn Rose, and Violet Peng, editors, \emph{Proceedings of the 2025 Conference on Empirical Methods in Natural Language Processing}, pages 20085--20095, Suzhou, China, November 2025. Association for Computational Linguistics.
\newblock ISBN 979-8-89176-332-6.
\newblock \doi{10.18653/v1/2025.emnlp-main.1015}.
\newblock \url{https://aclanthology.org/2025.emnlp-main.1015/}.

\end{thebibliography}

\clearpage
\newpage
\appendix

\section{Additional Experimental Results}

\subsection{Easy-to-Memorize Examples on Additional Models and Datasets}
\label{subsec:appendix_why_some_examples_easy_memorize}
\begin{figure}[H]
    \centering
    
    \begin{subfigure}[t]{0.48\columnwidth}
        \centering
        \includegraphics[width=\linewidth]{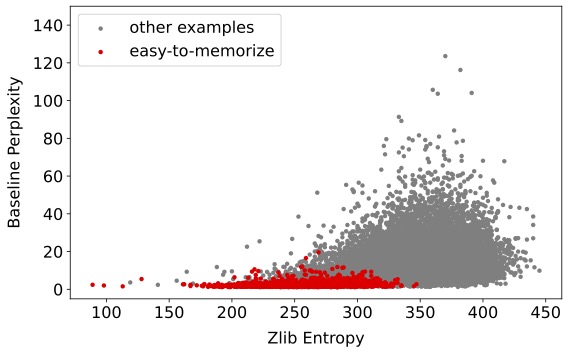}
        \caption{Pythia on Wikitext}
        \label{fig:pythia_wikitext}
    \end{subfigure}
    
    \vspace{0.3cm}

    \begin{subfigure}[t]{0.48\columnwidth}
        \centering
        \includegraphics[width=\linewidth]{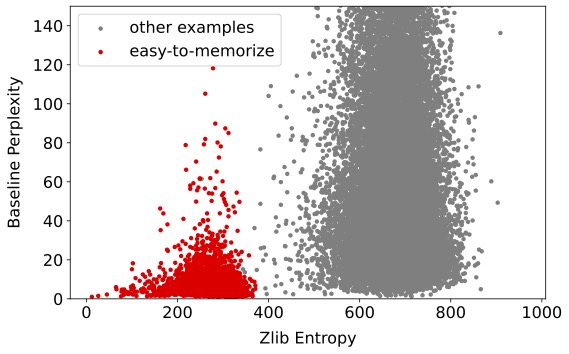}
        \caption{OLMo-2 on FineWeb}
        \label{fig:easy_olmo}
    \end{subfigure}
    \hfill
    \begin{subfigure}[t]{0.48\columnwidth}
        \centering
        \includegraphics[width=\linewidth]{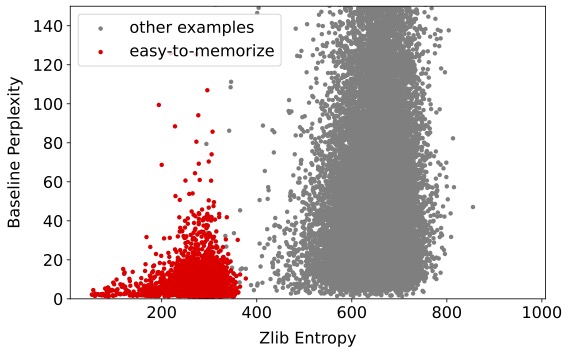}
        \caption{Qwen-3 on FineWeb}
        \label{fig:easy_qwen}
    \end{subfigure}

    \caption{\textbf{Intrinsic properties of \emph{easy-to-memorize examples}}. We plot the zlib entropy versus Baseline Perplexity for the \emph{easy-to-memorize} examples (red) compared to a random subset of 25,000 other training examples (grey). They form a distinct cluster with significantly lower entropy and perplexity.}
    \label{fig:zlib_vs_ppl}
\end{figure}

Figure~\ref{fig:zlib_vs_ppl} shows zlib entropy versus Baseline perplexity for \emph{easy-to-memorize} and other training examples across additional models and datasets: Pythia trained on Wikitext, and OLMo-2 and Qwen-3 trained on FineWeb. We observe similar clustering to that in Section~\ref{subsec:easy_to_memorize} (Figure~\ref{fig:easy_to_mem}), indicating that our findings generalize across models and datasets.

\subsection{Robustness to the memorization definition}
\label{subsec:mem_definition_robustness}
We check that the reduction in memorization from distillation holds under looser and stricter notions of memorization than our main exact 50-token definition.

For approximate memorization, we compute a BLEU score between the 50-token ground-truth suffix and the model's 50-token generation. Following~\citet{ippolito-etal-2023-preventing}, we consider a sequence approximately memorized if its BLEU score is greater than 0.75 and less than 1.0 (scores of exactly 1.0 correspond to exact matches). The fine-tuning baseline shows approximate memorization for 1,104 sequences, whereas the distilled student shows approximate memorization for 502 sequences, a reduction of 54.5\%. These findings are consistent with our exact-match evaluations, where the baseline memorizes 0.17\% and the student memorizes 0.07\% of training examples (Table~\ref{tab:mem_stats}), corresponding to a reduction of 58.8\%.

For the longer-prefix evaluation, we use a 100-token prefix and evaluate whether the model's 100-token generation exactly matches the ground-truth suffix. The fine-tuning baseline memorizes 630 examples, whereas the student memorizes 316 examples, consistent with our main findings. Together, these results confirm that distillation reduces memorization under both approximate and longer-prefix evaluation settings, strengthening our privacy claims.

\subsection{Extended Generalization Results}
\label{section:extended_generalization_results} 
Table~\ref{tab:extended_generalization_results} reports validation loss and perplexity for the Pythia family on Wikitext.

\begin{table}[h]
    \centering
    \caption{\textbf{Performance comparison on Wikitext.} Validation loss and perplexity for Pythia trained on Wikitext. The distilled student consistently outperforms the baseline.}
    \label{tab:extended_generalization_results} 
    \small
    \begin{tabular}{lcc}
        \toprule
        \textbf{Model} & \textbf{Val Loss} & \textbf{PPL} \\
        \midrule
        \multicolumn{3}{l}{\textit{Pythia Family (Wikitext)}} \\
        \hspace{3mm}Teacher (12B)   & 2.72 & 14.49 \\
        \hspace{3mm}Baseline (1.4B) & 2.82 & 16.33 \\
        \hspace{3mm}Student (1.4B)  & \textbf{2.75} & \textbf{15.36} \\
        \bottomrule
    \end{tabular}
\end{table}

\subsection{Extended Analysis: Other Models and Datasets}
\label{sec:extended_analysis_models_datasets}
In this section, we extend our analysis to Pythia models trained on the Wikitext dataset, Olmo-2 models trained on FineWeb, and Qwen-3 models trained on FineWeb. 

\subsubsection{Memorization Overlap Analysis}
\label{sec:extended_mem_overlap_analysis}   
\begin{figure}[H]
    \centering
    \begin{subfigure}[t]{0.3\textwidth}
        \centering
        \includegraphics[width=\linewidth]{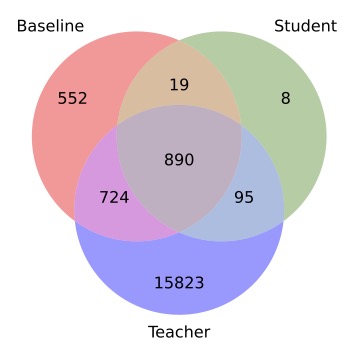}
        \caption{Pythia}
        \label{fig:pythia}
    \end{subfigure}
    \hfill
    \begin{subfigure}[t]{0.3\textwidth}
        \centering
        \includegraphics[width=\linewidth]{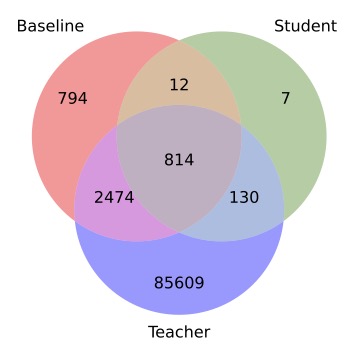}
        \caption{OLMo-2}
        \label{fig:olmo}
    \end{subfigure}
    \hfill
    \begin{subfigure}[t]{0.3\textwidth}
        \centering
        \includegraphics[width=\linewidth]{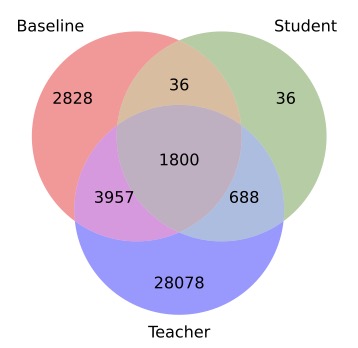}
        \caption{Qwen-3}
        \label{fig:qwen}
    \end{subfigure}

    \caption{\textbf{Memorization overlap across architectures and datasets.} Venn diagrams visualizing the intersection of memorized examples between Teacher, Baseline, and Student models for \textbf{(a)} Pythia 1.4B on Wikitext, \textbf{(b)} OLMo-2 1B on FineWeb, and \textbf{(c)} Qwen-3 1.7B on FineWeb. Across all settings, the Student preferentially memorizes \emph{easy to memorize} examples (examples that both the Teacher and Baseline memorize) while showing some inheritance (blue region) from the Teacher.}
    \label{fig:overlap_extended}
\end{figure}

Figure~\ref{fig:overlap_extended} illustrates the memorization overlap between the Teacher, Baseline, and Student models for Pythia (trained on Wikitext), OLMo-2 (FineWeb), and Qwen-3 (FineWeb). Consistent with the findings in Section~\ref{subsec:easy_to_memorize}, we observe that Students preferentially memorize \emph{easy-to-memorize} examples, exhibiting significant overlap with both the Teacher and Baseline. Specifically, examples categorized as easy-to-memorize account for 88\% of the Pythia Student's total memorization, 85\% for the OLMo-2 Student, and 70\% for the Qwen-3 Student. In terms of inheritance (shown by blue region), the Pythia Student inherits 9\% of its memorization directly from the Teacher, while the OLMo-2 and Qwen-3 Students inherit 13\% and 27\%, respectively.

\subsubsection{Feature Distributions Distinguishing Memorized vs. Non-Memorized Examples}
\label{sec:extended_all_features} 
\begin{figure*}[h]
    \centering
    \includegraphics[width=0.9\textwidth]{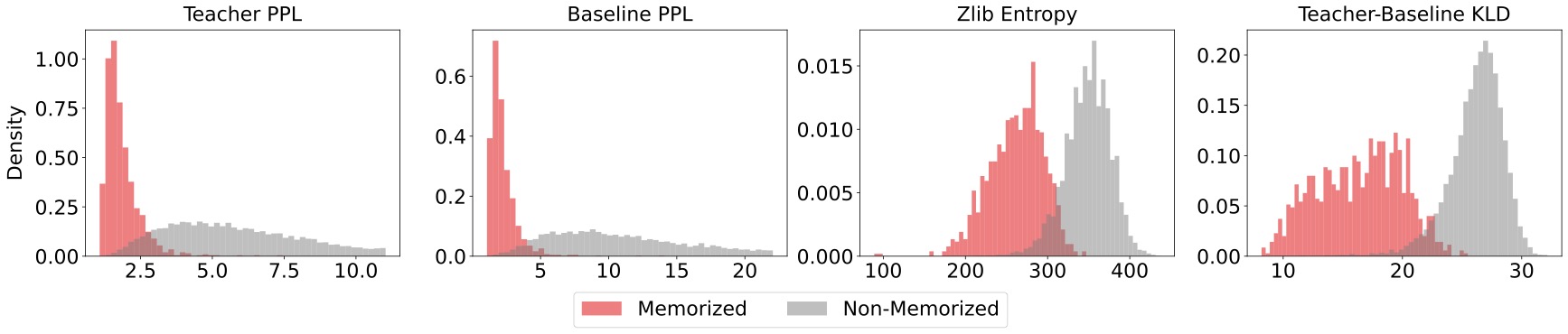}  
    \caption{\textbf{Feature distributions distinguishing memorized vs. non-memorized examples in Student (Pythia on Wikitext)}. Teacher Perplexity is the most significant predictor, followed by zlib Entropy, the KL divergence loss between the Teacher and Baseline, and Baseline Perplexity. Across all metrics, lower values are consistently associated with memorized examples.}
    \label{fig:all_metrics_pythia_wikitext}
\end{figure*}

\begin{figure*}[h]
    \centering
    \includegraphics[width=0.9\textwidth]{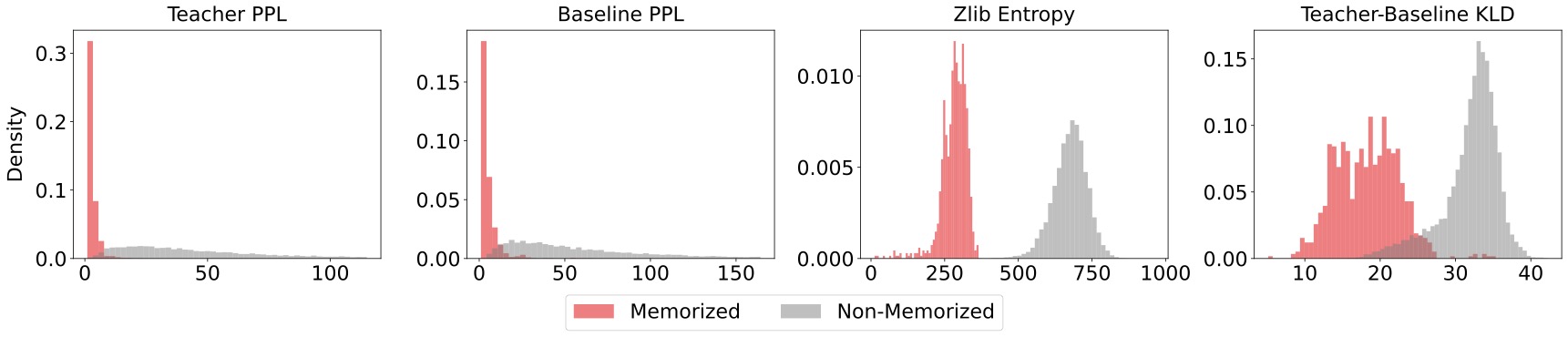}  
    \caption{\textbf{Feature distributions distinguishing memorized vs. non-memorized examples in Student (OLMo-2 on FineWeb)}. zlib Entropy is the most significant predictor, followed by the KL divergence loss between the Teacher and Baseline, Teacher Perplexity, and Baseline Perplexity. Across all metrics, lower values are consistently associated with memorized examples.}
    \label{fig:all_metrics_olmo2}
\end{figure*}

\begin{figure*}[h]
    \centering
    \includegraphics[width=0.9\textwidth]{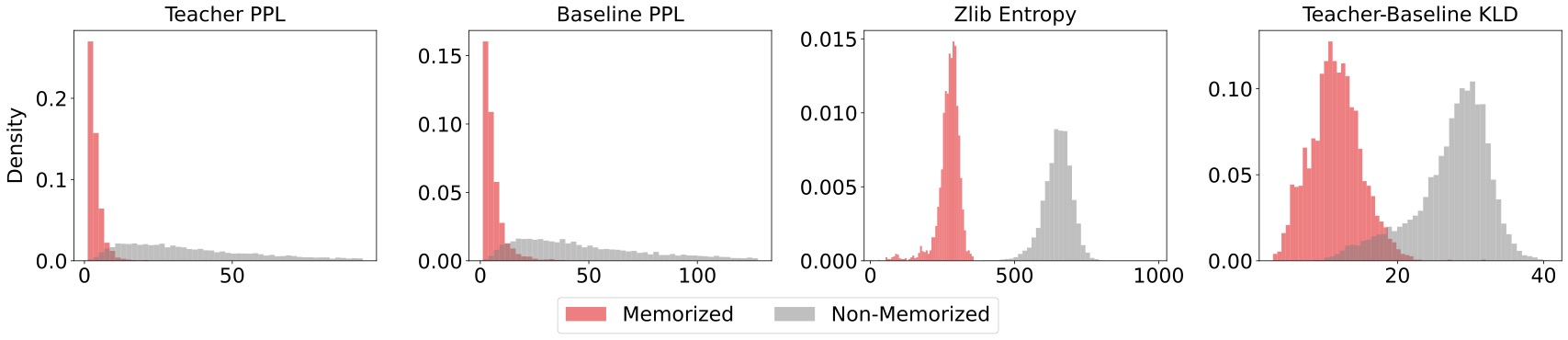}  
    \caption{\textbf{Feature distributions distinguishing memorized vs. non-memorized examples in Student (Qwen-3 on FineWeb)}. zlib Entropy is the most significant predictor, followed by the KL divergence loss between the Teacher and Baseline, Teacher Perplexity, and Baseline Perplexity. Across all metrics, lower values are consistently associated with memorized examples.}
    \label{fig:all_metrics_qwen3}
\end{figure*}

Figure~\ref{fig:all_metrics_pythia_wikitext} shows the distribution of features used to train the memorization classifier for the Pythia model on Wikitext. Figure~\ref{fig:all_metrics_olmo2} presents the corresponding distributions for the OLMo-2 model trained on FineWeb, and Figure~\ref{fig:all_metrics_qwen3} for the Qwen-3 model trained on FineWeb. 

\subsection{More Details On Our Memorization Classifier} 
\label{sec:extended_mem_classifier}
Table~\ref{tab:feature_importance} reports the coefficients learned by our logistic regression classifier for predicting memorization in the Pythia student distilled on FineWeb. Table~\ref{tab:feature_importance_pythia_wikitext} presents the corresponding results for Pythia trained on Wikitext, while Tables~\ref{tab:feature_importance_olmo2} and~\ref{tab:feature_importance_qwen3} report results for OLMo-2 and Qwen-3 on FineWeb, respectively.

\begin{table}[H]
    \centering
    \small   
    \setlength{\tabcolsep}{2pt}  
    \caption{Feature Importance based on standardized Logistic Regression coefficients (averaged over 100 trials). The large negative magnitude of zlib Entropy indicates it is the strongest predictor. \textbf{Results are for Pythia models trained on FineWeb.}}
    \label{tab:feature_importance}
    \begin{tabular}{lc}
        \toprule
        \textbf{Feature} & \textbf{Coeff. (Mean $\pm$ Std)} \\ 
        \midrule
        Teacher PPL           & $-0.3341 \pm 0.1228$ \\
        Baseline PPL          & $-0.4010 \pm 0.1417$ \\
        zlib Entropy          & $\mathbf{-4.5001 \pm 0.1324}$ \\
        Teacher–Baseline KLD  & $-1.0579 \pm 0.1664$ \\
        \bottomrule
    \end{tabular}
\end{table}

\begin{table}[h]
\centering
\caption{Memorization classifier results for \textbf{Pythia} trained on \textbf{Wikitext}. We report mean $\pm$ standard deviation over 100 trials. Feature importance corresponds to standardized logistic regression coefficients.}
\label{tab:feature_importance_pythia_wikitext}
\begin{tabular}{lcc}
\toprule
\textbf{Metric} & \textbf{Value} & \\
\midrule
Accuracy  & $0.9799 \pm 0.0038$ \\
Precision & $0.9515 \pm 0.0111$ \\
Recall    & $0.9690 \pm 0.0095$ \\
F1        & $0.9601 \pm 0.0075$ \\
ROC-AUC   & $0.9964 \pm 0.0010$ \\
\midrule
\textbf{Feature} & \textbf{Coefficient} & \\
\midrule
Teacher PPL          & $-2.7531 \pm 0.2852$ \\
Baseline PPL         & $-1.9694 \pm 0.2530$ \\
zlib Entropy         & $-2.0809 \pm 0.1712$ \\
Teacher--Baseline KLD& $-2.0074 \pm 0.2294$ \\
\bottomrule
\end{tabular}
\end{table}

\begin{table}[h]
\centering
\caption{Memorization classifier results for \textbf{OLMo-2} trained on \textbf{FineWeb}. We report mean $\pm$ standard deviation over 100 trials. Feature importance corresponds to standardized logistic regression coefficients.}
\label{tab:feature_importance_olmo2}
\begin{tabular}{lcc}
\toprule
\textbf{Metric} & \textbf{Value} & \\
\midrule
Accuracy  & $0.9988 \pm 0.0010$ \\
Precision & $0.9952 \pm 0.0040$ \\
Recall    & $1.0000 \pm 0.0000$ \\
F1        & $0.9976 \pm 0.0020$ \\
ROC-AUC   & $0.9999 \pm 0.0002$ \\
\midrule
\textbf{Feature} & \textbf{Coefficient} & \\
\midrule
Teacher PPL          & $-0.1151 \pm 0.0837$ \\
Baseline PPL         & $-0.1854 \pm 0.0877$ \\
zlib Entropy         & $-4.4637 \pm 0.1376$ \\
Teacher--Baseline KLD& $-1.5805 \pm 0.1455$ \\
\bottomrule
\end{tabular}
\end{table}

\begin{table}[h]
\centering
\caption{Memorization classifier results for \textbf{Qwen-3} trained on \textbf{FineWeb}. We report mean $\pm$ standard deviation over 100 trials. Feature importance corresponds to standardized logistic regression coefficients.}
\label{tab:feature_importance_qwen3}
\begin{tabular}{lcc}
\toprule
\textbf{Metric} & \textbf{Value} & \\
\midrule
Accuracy  & $0.9991 \pm 0.0004$ \\
Precision & $0.9963 \pm 0.0017$ \\
Recall    & $1.0000 \pm 0.0000$ \\
F1        & $0.9981 \pm 0.0009$ \\
ROC-AUC   & $0.9999 \pm 0.0002$ \\
\midrule
\textbf{Feature} & \textbf{Coefficient} & \\
\midrule
Teacher PPL          & $-0.1576 \pm 0.1053$ \\
Baseline PPL         & $-0.1184 \pm 0.0710$ \\
zlib Entropy         & $-5.4524 \pm 0.1125$ \\
Teacher--Baseline KLD& $-1.5311 \pm 0.1340$ \\
\bottomrule
\end{tabular}
\end{table}

\newpage
\subsection{Dissecting Student and Baseline Memorization}
Next, we are interested in whether it is possible to identify which examples would be memorized by the student model during distillation and which would be memorized by the baseline model when trained from scratch. From the results in Section~\ref{sec:memorize_less}, we know that distilled student models memorize less than the baseline, even though both models have a similar number of parameters. From Section~\ref{subsec:easy_to_memorize}, we see that the majority of examples memorized by the student are also memorized by the baseline. However, the baseline model memorizes many additional examples that are never memorized by the student. This raises the question: are there factors that can predict which examples will be memorized by the student and which will be memorized \emph{exclusively} by the baseline model?

Let there be two datasets of memorized examples, where $D_s$ represents examples memorized by the student, and $D_b$ represents examples memorized by the baseline but not student. We mount a membership inference attack (MIA)~\citep{MIA_OG} using $D_s$ \& $D_b$ as datasets and compute the area under the curve (AUC) using the following features: teacher perplexity, baseline perplexity, and the KLD loss between teacher and student. We observe an AUC of 0.745 for teacher perplexity, 0.754 for baseline perplexity, and 0.855 for the KLD loss between teacher and student. This shows that these metrics are good indicators of which examples are memorized by the student and the baseline.

\begin{figure}[H]
 \centering
  \includegraphics[width=0.8\columnwidth]{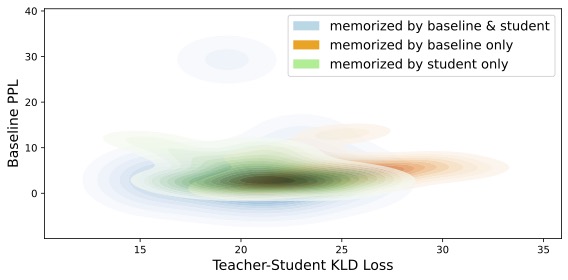}
  \caption{Kernel Density Estimation (KDE) plot showing probability distribution of baseline perplexity and teacher-student KLD loss across three datasets: examples memorized by baseline \& student, examples memorized by baseline-only, and examples memorized by student-only. Results are for Pythia models trained on the FineWeb.}
  \label{fig:d1_d2_de}
\end{figure}

Figure~\ref{fig:d1_d2_de} displays Kernel Density Estimation (KDE) contours correlating baseline perplexity with teacher-student KLD loss across three categories of memorized examples.

The blue region represents easy-to-memorize examples that are memorized both by baseline and student models. These examples typically have lower baseline perplexity and teacher-student KLD loss. However, we do see a  tail extending to higher PPL values. This suggests that this group does contain a subset of examples that were inherently harder for the baseline to memorize. 

The orange region shows examples that are memorized exclusively by the baseline model. This cluster is tightly concentrated in the region of lowest Baseline PPL ($<10$) and highest teacher-student KLD loss indicating these examples were never memorized by the student. 

The green region highlights examples memorized exclusively by the student model. These examples appear to be concentrated in the low KLD-loss region and, interestingly, also have lower baseline perplexity. This may indicate that these examples are on the verge of being memorized by the baseline model.

\subsection{Evaluating Memorization During Distillation in a Pre-training Setup}
\label{subsec:pretrain_distillation}
\begin{figure}[H]
 \centering
  \includegraphics[width=0.6\columnwidth]{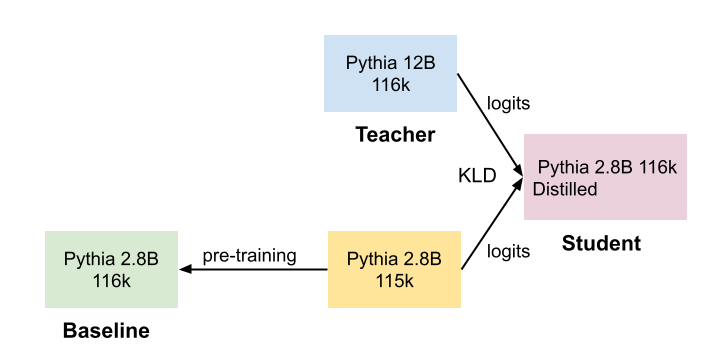}
  \caption{Our Knowledge Distillation Framework in a Pre-training Setup for Pythia family models.}
  \label{fig:pretrain_kd}
\end{figure}

We also ran some early experiments to understand the dynamics of memorization for logit-level distillation in a pre-training setup (Figure~\ref{fig:pretrain_kd}) using the formulation from equation~\ref{eqn:eqn1}. We use the 116k checkpoint of Pythia 12B as the Teacher. We initialize the Student from the 115k checkpoint of Pythia 2.8B and continue training it on the data seen by the Teacher between checkpoints 115k and 116k ($\approx$ 1M examples) using KL divergence ($T = 2$) for five epochs. We also use the 116k checkpoint of Pythia 2.8B from the Pythia suite as a Baseline. Note that all Pythia family models see exactly the same data between each checkpoint during training. We then compare memorization rates across the Teacher, Baseline, and Student. 

Our results correlate strongly with memorization trends observed in our main fine-tuning KD setup. As shown in Table~\ref{tab:memorization_pretrain}, the Teacher has a memorization rate of 1.57\%, the Baseline has 0.27\%, and the Student has a rate of \textbf{0.06\%}. Of 2785 examples memorized by the Baseline, 2117 (76\%) are also memorized by the teacher, indicating that the \emph{easy-to-memorize} examples are prevalent in this setup as well. Among these 2,117 examples, only 361 are memorized by the student, highlighting the regularization effect of distillation. Finally, we identify 92 examples that are memorized by the student but never memorized by either the teacher or the baseline.

\begin{table}[h]
    \centering
    \caption{\textbf{Memorization statistics.} Comparison of memorization across the Teacher, Baseline, and Student models for approximately 1 million training examples. The Student model exhibits significantly lower memorization compared to the Baseline.}
    \label{tab:memorization_pretrain}
    \small
    \begin{tabular}{lcc}
        \toprule
        \textbf{Model} & \textbf{\# Examples Memorized} & \textbf{Memorization \%} \\
        \midrule
        Teacher (12B)   & 16,133 & 1.57\% \\
        Baseline (2.8B) & 2,785  & 0.27\% \\
        Student (2.8B)  & \textbf{638}    & \textbf{0.06\%} \\
        \bottomrule
    \end{tabular}
\end{table}

\end{document}